\documentclass[journal]{IEEEtran}

\usepackage[noadjust]{cite}
\usepackage{graphicx}
\usepackage{amsmath}
\usepackage{amssymb}
\usepackage{multirow}
\usepackage[letterpaper=true,colorlinks,bookmarks=false]{hyperref}
\usepackage{subfigure}
\usepackage[lined,boxed, ruled]{algorithm2e}
\title{  Structure-preserving Guided Retinal Image Filtering and Its Application for Optic Disc Analysis}

\author{Jun Cheng*, Zhengguo Li, Zaiwang Gu, Huazhu Fu, Damon Wing Kee Wong,  Jiang Liu 
\thanks{
	Manuscript received March 19, 2018; revised May 12; accepted May 16, 2018. This work was supported in part under grant Y80002RA01 by Cixi institute of biomedical engineering, Chinese Academy of Sciences, China and Ningbo 3315 Innovation
	team grant Y61102DL03. \emph{Asterisk indicates corresponding author}.}
\thanks{*J. Cheng is with Cixi Institute of Biomedical Engineering, Chinese Academy of Sciences, China. (email: sam.j.cheng@gmail.com, chengjun@nimte.ac.cn)}
 \thanks{Z. Li is  with Institute for Infocomm Research,  Agency for Science, Technology
 	and Research, Singapore. (email: ezgli@i2r.a-star.edu.sg).   .
}
  \thanks {Z. Gu is with Cixi Institute of Biomedical Engineering, Chinese Academy of Sciences, and Shanghai University, China (email: guzaiwang@nitme.ac.cn).
  }
 \thanks{H. Fu and D. W. K. Wong are with Ocular Imaging (iMED) Department in  Institute for Infocomm Research, Agency for Science, Technology
 and Research, Singapore (email: huazhufu@gmail.com, wkwong@i2r.a-star.edu.sg).}
 
  \thanks{J. Liu is with Cixi Institute of Biomedical Engineering,
  Chinese Academy of Sciences, China (email: jimmyliu@nimtec.ac.cn)}

      }
\begin{document}
 \markboth{IEEE TRANSACTIONS ON MEDICAL IMAGING,~Vol.~x, No.~x,
~2018}
  {Cheng \MakeLowercase{\textit{et al.}}:}

\maketitle
\begin{abstract} Retinal fundus photographs have been used in the  diagnosis of many  ocular diseases such as  glaucoma, pathological myopia, age-related macular degeneration and diabetic retinopathy. With the development of computer science, computer aided diagnosis has been developed to process and analyse the retinal images automatically.  One of the challenges in the analysis is that the  quality of the retinal image is often degraded. For example, a cataract in human lens will attenuate the retinal image, just as a cloudy camera lens which reduces the quality of
a photograph. It often obscures the details in the retinal images and
posts   challenges in retinal image processing and analysing tasks. In this paper, we  approximate the degradation  of the retinal images as a combination of human-lens attenuation and scattering. A novel  structure-preserving guided retinal image filtering (SGRIF) is then proposed to restore images based on the  attenuation and scattering model. The proposed SGRIF consists of a step  of global structure transferring and a step of global edge-preserving smoothing. Our results show that the proposed SGRIF method is able to improve the contrast of retinal images, measured by histogram flatness measure, histogram spread and variability of local luminosity. In addition, we further explored the benefits of  SGRIF for subsequent retinal image processing and  analysing tasks. In the two applications of deep learning based optic cup segmentation and sparse learning based cup-to-disc ratio (CDR) computation, our results show that we are able to achieve more accurate optic cup segmentation  and CDR measurements from images processed by SGRIF.

\end{abstract}
\section{Introduction} 
Retinal fundus photographs have been widely used by clinicians to diagnose and monitor many ocular diseases including glaucoma,  pathological myopia, age-related macular degeneration, and diabetic retinopathy. Since manual assessment of the images is tedious, expensive and subjective, computer aided diagnosis methods \cite{abramoff10} have been developed to analyse the images automatically for optic disc segmentation \cite{huiqi03, aquino2010, Muramatsu11, jcheng2011},  optic cup segmentation or cup to disc ratio assessment  \cite{ abramoff07, 6464593, CJ15, huazhu18},  retinal vessel detection \cite{Drive,8049478},  glaucoma detection \cite{annan16,  Mishra13},  diabetic retinopathy detection \cite{AbramoffDR, doi:10.1001/jama.2016.17216},  age-related macular degeneration detection  \cite{KOSE2008611, 5627289, doi:10.1080/10255842.2011.623677, doi:10.1167/iovs.10-7075}, and pathological myopia detection \cite{JHEPPA}.  Image quality is an essential factor for proper development and validation of the algorithms \cite{Trucco13}. Very often, low quality images lead to poor performance of  automatic analysis.  Therefore, it is necessary to restore the image for better analysis.  There are many factors that may affect the quality of retinal images.  In retinal imaging, an illumination light  passes through the lens of   human eye to reach the retina. Ideally, the light will  be reflected back to the fundus camera by the retina to form the retinal images. However, the human eye is not a perfect optical system and the   light received by the fundus camera is often attenuated along the   path of the light.  This can be serious when the lens of   human eye is affected by  diseases  such as cataracts. 

\begin{figure}
	\centering
	{\subfigure[]{\label{fig1b}
			\includegraphics[height=1.1in ]{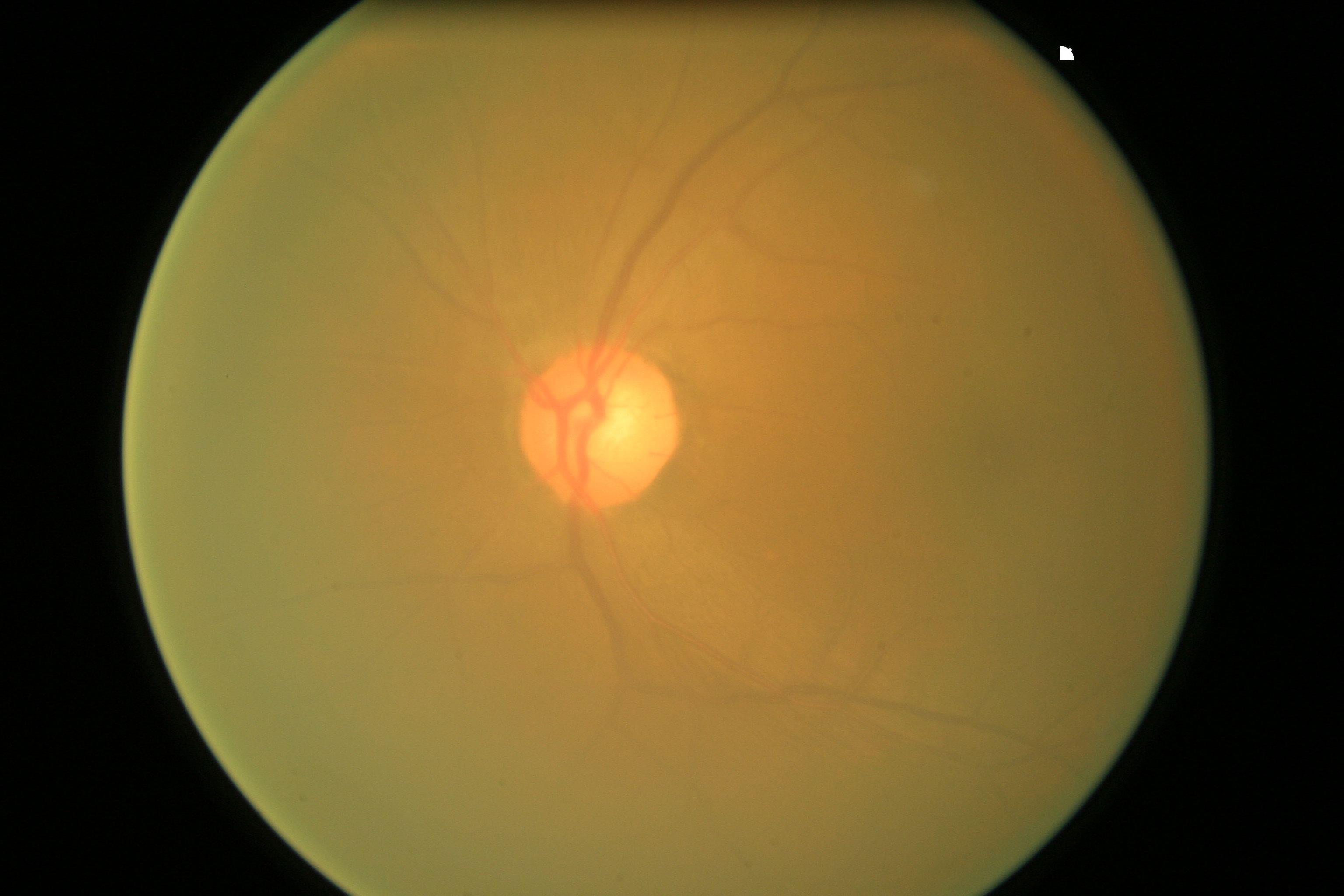}
	}}\hspace{-0.25cm}
	{\subfigure[]{\label{fig1c}
			\includegraphics[height=1.1in ]{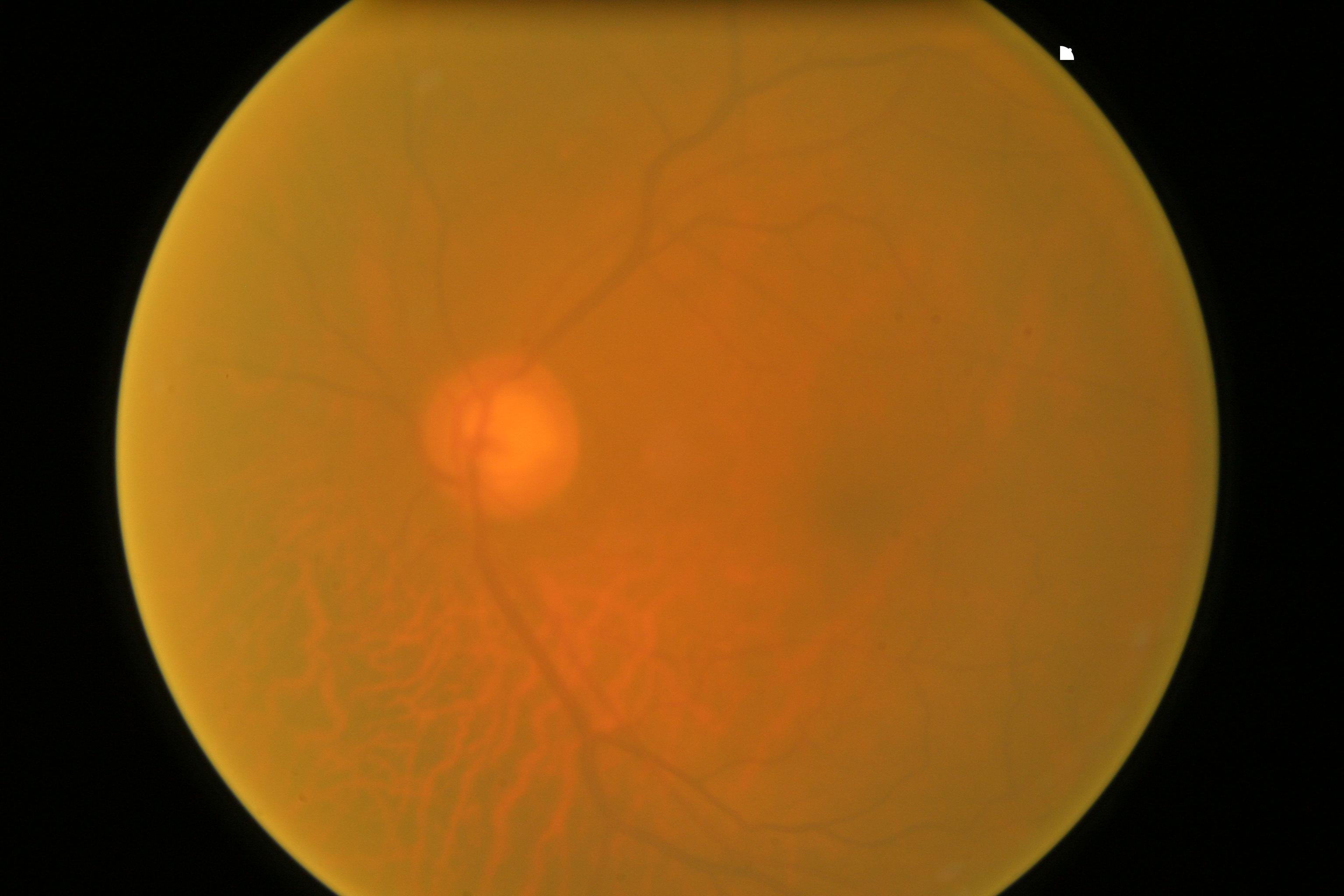}
	}}  
	{\subfigure[]{\label{fig1d}
			\includegraphics[height=1.1in ]{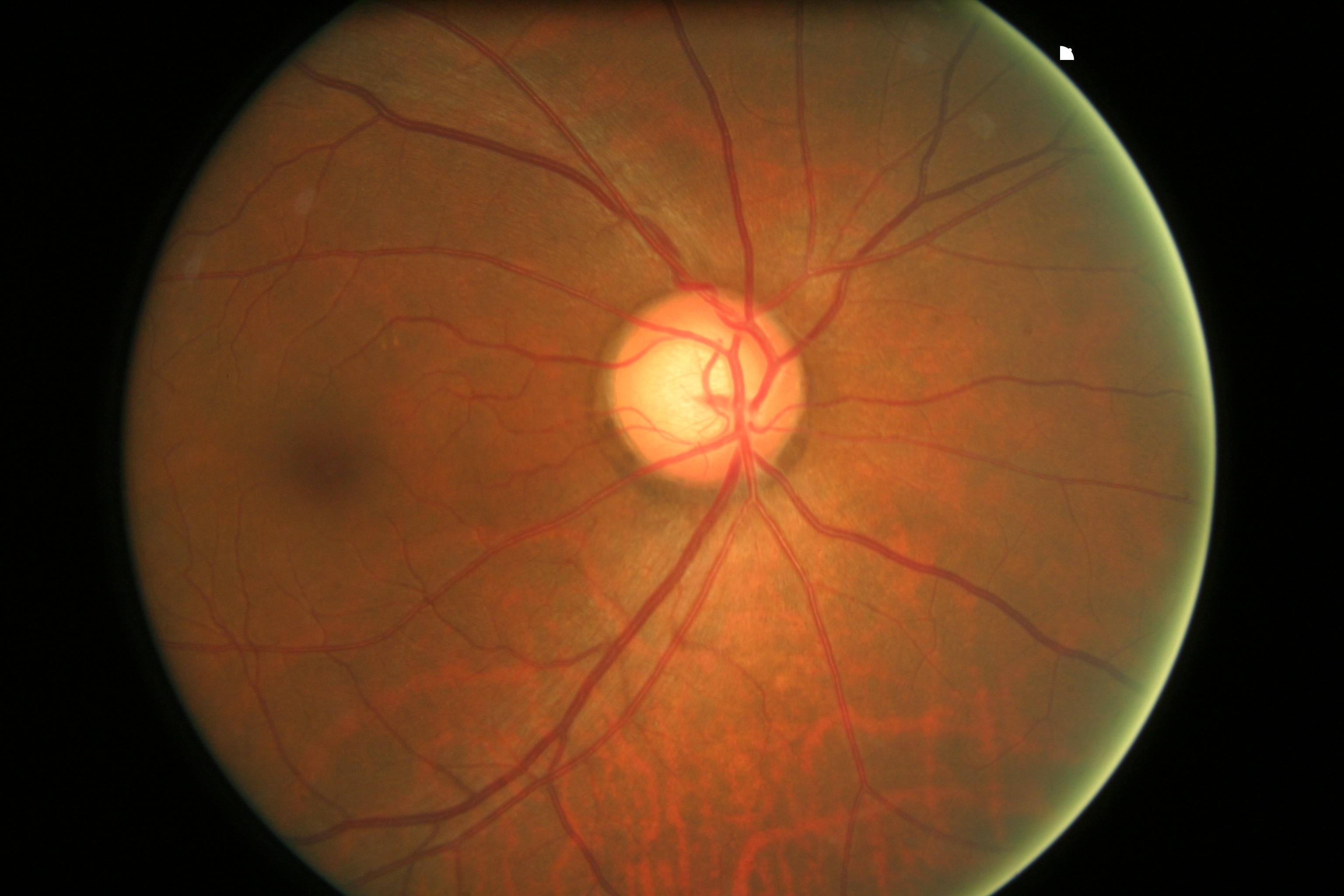}
	}}\hspace{-0.25cm}
	{\subfigure[]{\label{fig1e}
			\includegraphics[height=1.1in ]{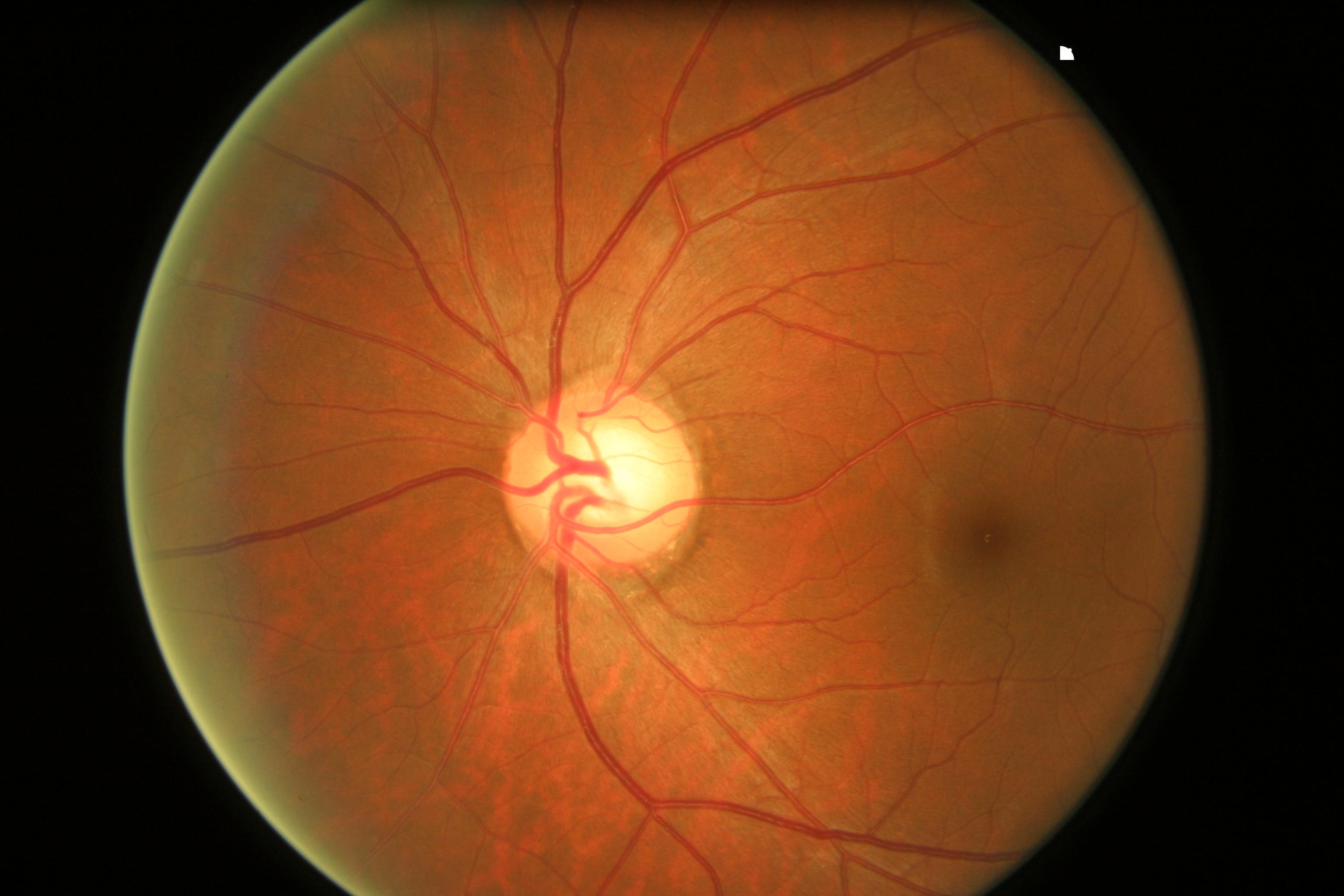}
	}}\\
	\caption{Retinal images: (a)-(b) from eyes with cataract and (c)-(d) from eyes without cataract } \label{fig1}
\end{figure}

 Cataract is caused by clouding of   human-lens in the eye.  Since a retinal image  is captured through the lens of human eye, a cataract  will lead to attenuation and scattering of the light travelling through the human-lens, just as a cloudy camera lens reducing the quality of a photograph.  The  process of retina images by cataractous human-lens is referred to as clouding and the processing to remove the effect is referred to as declouding.  Meanwhile, cataract  accounts for 33\%  of blindness worldwide \cite{Pascolini614} and its  global prevalence  in adults over 50 years of age was about 47.8\% in 2002 \cite{Resnikoff04}. The high prevalence of  cataract makes it an innegligible factor that affects retinal imaging. 
Depending on the severity and the location of the clouding in the  cataractous lens, the
retinal images may be degraded at different levels of severity.  It often
obscures the details of the images, affects the diagnostic process and   posts  challenges in
retinal image processing and analysis.   Fig. \ref{fig1} shows four samples  
of retinal images, where the first row shows two images from eyes with 
cataract and the second row shows two images from  eyes without any cataract.
As we can see, the cataract degrades the retinal images and reduces
the dynamic range of the images.  Since the degradation is caused by light scattered by the human-lens, we call the scattered light  as lens-light.     Removal of the cloudy effect introduced by lens-light will increase the contrasts of the retinal image  and correct the  distortion. 

Besides cataracts, many other factors affect the  quality of retinal images as well.
Studies have shown that refractive anomalies such as
defocus or astigmatism often  blur the image \cite{Marcos03}. Optical aberration   also blurs  the retinal image, even though its contribution is  typically smaller than   that of defocus or astigmatism \cite{Liang:97}. Studies have shown there is a large increment of ocular aberration with age \cite{Artal93}. Scattering in normal eyes is usually small. However, it is also known that the scattering increases with age.   These different factors make the visual appearance of the  retinal images from different subject eyes different.

In retinal image processing,  the low image quality due to the clouding often results in many challenges to learn a good representation of the images for  structure detection or segmentation, lesion detection and other analysis. In optic cup segmentation, the boundary between the optic cup and neuroretinal rim may have a wide range of gradients and may have been affected by cataractous lens or other factors. In haemorrhage detection or vessel segmentation, the intensity of the vessels may vary largely from one image to another and the vessel boundaries might also be obscured. This motivates us to remove the clouding  effect due to the scattering of human-lens and increase the contrast of the retinal images.

 In  \cite{Peli1989},  the degradation  due to cataracts is modelled as 
 \begin{equation}
 I_c(p)=  \alpha L_cr_c(p)t(p)+L_c(1-t(p)), \label{eq1}
 \end{equation}
 where $\alpha$ is a constant denoting  the attenuation
 of retinal illumination due to the cataracts, $c\in \{r,g,b\}$ denotes the red, green or blue channel of the image,  $r_c(p)$ denotes the retinal reflectance function, $L_c$ is the  illumination of the fundus camera and $t(p)$ describes the portion of the light that does not reach the camera. 
 
 The above model requires a precataract clear image to estimate  $\alpha$. However, such image is often unavailable and the illumination light may alse be different if the precataract image is caputured  under different conditions.  On the other hand, the value of the constant $\alpha$ only affects the final output image linearly. One can apply a linear scale $1/\alpha$ on the output image if a precataract image is available to determine $\alpha$.  	
 	 Therefore, we propose to use a slightly  simpler model: 
     \begin{equation}
	 I_c(p)=  D_c(p)t(p)+L_c(1-t(p)), \label{eq111}
	 \end{equation}
	 where $D_c(p)= L_cr_c(p)$  denotes the image captured under
	 ideal human-lens. 
	The model in equation (\ref{eq111}) is the same as the 
  dehaze model in computer vision where the attenuation of the images by the haze or fog are modeled by air attenuation and scattering  \cite{Tan2008VisibilityIB}.
  This model can be considered as a special case of  (\ref{eq1}) by letting  $\alpha=1$, i.e., ignoring the attenuation $\alpha$.  
 By applying the model on the retinal image, the problem of removing the clouding effect due to lens scattering becomes a common dehaze problem in computer vision.  

  In computer vision, many methods \cite{Tan2008VisibilityIB, Fattal08, Hekaiming2011, Hekaiming2013, li2015wgif, li2015edge} have been proposed to solve the  dehaze problem in equation (\ref{eq111}).   In \cite{Tan2008VisibilityIB},  Markov random field was used to enhance the local contrast of the restored image, however, it often produces
 over-saturated images.  
  In \cite{Fattal08}, Fattal proposed a model  that accounts for both surface shading and scene transmission, but it does not perform well in heavy haze.   In \cite{Hekaiming2011}, a novel dark channel prior was used.  But this assumption is not always valid for retinal images which do not have many shadows or complex structures as natural scene images. Guided image filtering (GIF) \cite{Hekaiming2013} is a promising technology  proposed recently for single image haze removal. It has a limitation that it often oversmoothes the regions close to flat and does not preserve the fine structure which might be important in retinal image processing or analysis. To overcome the limitation, we propose a method to preserve the structure in the original images.
Motivated by GIF, we propose a structure-preserving guided retinal image filtering (SGRIF), which is composed of a global structure transfer filter and a global edge-preserving smoothing filter. 
Then we apply SGRIF on  the retina images for  subsequent analysis.   Different from most of work that ends with image quality evaluation, we further investigate how the process benefits subsequent automatic analysis from the processed images.  Two different applications including  deep learning based optic disc segmentation and sparse-learning based cup-to-disc ratio (CDR) computation are conducted to show the advantage of our method. 

\textbf{Contribution:} Our main contributions are summarized as
follows.  
\begin{enumerate}
	\item 
 We propose  a structure preserving guided retinal image filtering   for declouding of the retinal images. 

\item Our experiments show  that the proposed SGRIF algorithm is able to improve  the contrast of the image and maintain  the edges for further analysis. 

\item  Our method benefits subsequent analysis as well.  In applications of deep learning based optic disc segmentation, the pre-processing using the proposed SGRIF improves the segmentation results. In application of sparse-learning based CDR computation, we also show that the proposed method improves the accuracy  of CDR computation.
\end{enumerate}

The remaining sections of the paper are organized as follows. Section \ref{spgrif}  introduces our proposed method to remove the clouding effect  caused by lens-light. The method includes a step of global structure transferring and a step of global edge-preserving smoothing. Experimental results are given in Section \ref{results} to show the effectiveness of the proposed method to improve the contrast of the retinal images and its application in optic cup segmentation and CDR computation. Conclusions are provided in the last section.

\section{Structure-preserving guided retinal image filtering} \label{spgrif}

In GIF \cite{Hekaiming2013}, a guidance image $G$  which could be identical to the input image $I$ is
used to guide the process. The output image $O$ is computed by a linear
transform of the guidance image $G$ in a window $W_p$ centered at the pixel $p$, i.e.,
  \begin{equation}
    O_i=a_p G_i+b_p,  \forall i\in W_p, \label{eq3}
 \end{equation}
 where $i$ indicates pixel index. 
 
 The linear transform coefficients $a_p, b_p$ are computed by minimizing the following objective function：
 \begin{equation}
    E=(O_i-I_i)^2+\epsilon a_p^2,  \label{eq4}
 \end{equation}
where   $\epsilon$ is a regularization parameter.

Its solution is given by:
\begin{equation}
   a_p=\frac{\frac{1}{|W_p|}\sum_{i\in W_p} G_iI_i- \mu_p \bar{I}_p}{\sigma_p^2+\epsilon}, \label{eq5}
\end{equation}
\begin{equation}
b_p=\bar{I}_p-a_p\mu_p,  \label{eq6}
\end{equation}
 where $\bar{I}_p$ denotes the mean of $I$ in $W_p$, $|W_p|$ is the cardinality of $W_p$, $\mu_P$ and $\sigma_p^2$ denote the mean and variance of $G$ in $W_p$ respectively.

Since each pixel $i$ is covered by many overlapping window $W_p$, the output of equation (\ref{eq5}) from different windows is not identical. An averaging strategy is used in GIF \cite{Hekaiming2013}, i.e.,
\begin{equation}
   O_i=\bar{a}_pG_i+\bar{b}_p,  \label{eq7}
\end{equation}
where $\bar{a}_p$ and $\bar{b}_p$ are the mean values of $a_{p'}$ and $b_{p'}$ in   $W_p$: 
\begin{equation}
    \bar{a}_p=\frac{1}{|W_p|}\sum_{p'\in {W_p}} a_{p'},
\end{equation}
\begin{equation}
\bar{b}_p=\frac{1}{|W_p|}\sum_{p'\in {W_p}} b_{p'}.
\end{equation}
 \begin{figure*}
	\centering
	{\subfigure[]{\label{fig11b}
			\includegraphics[height=2.2in]{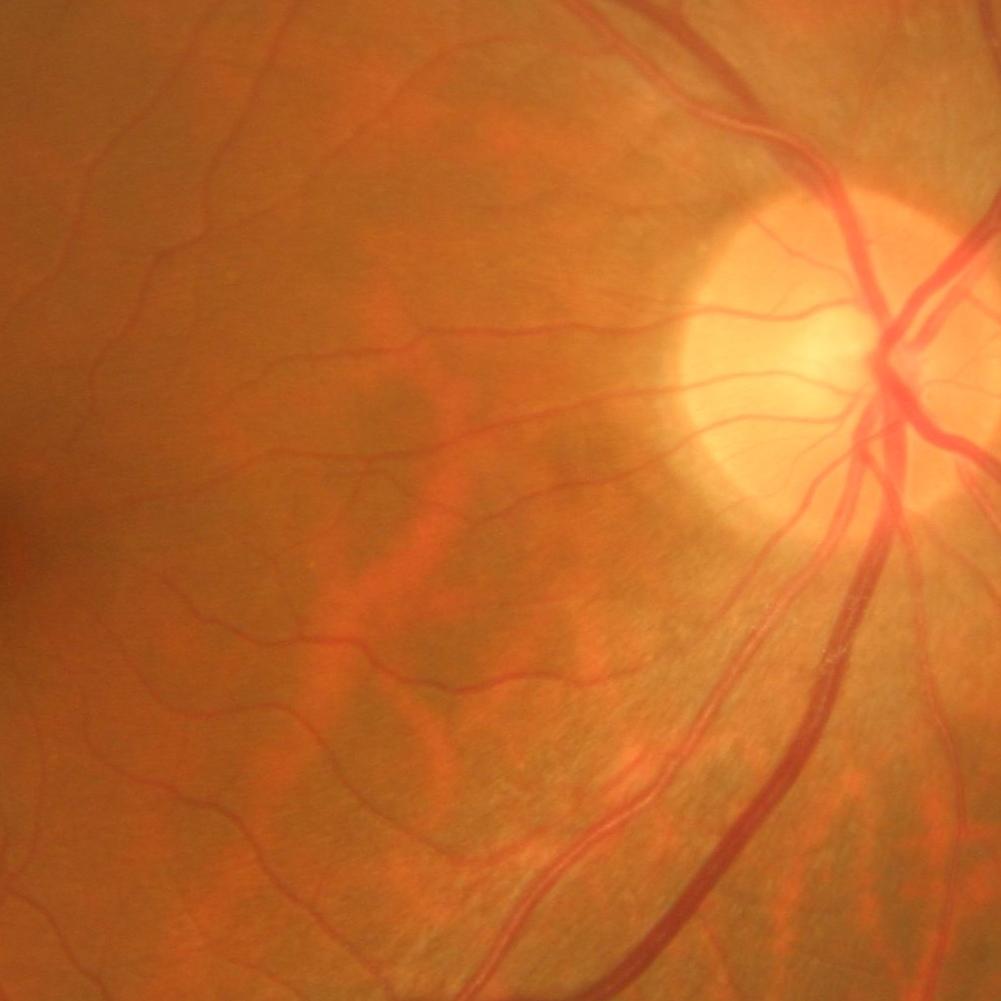} \label{figneeda}
	}} 
	{\subfigure[]{\label{fig11c}
			\includegraphics[height=2.2in]{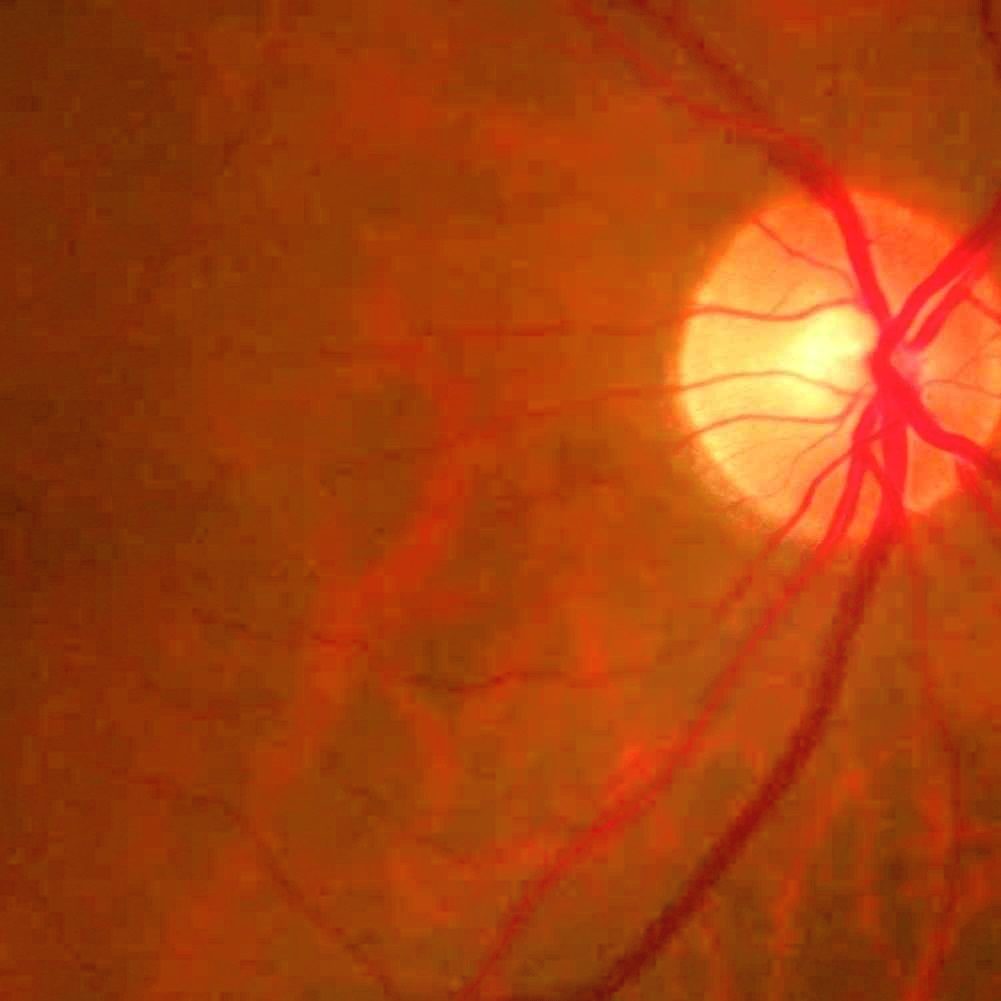}
	}} 
	{\subfigure[]{\label{fig11d}
			\includegraphics[height=2.2in]{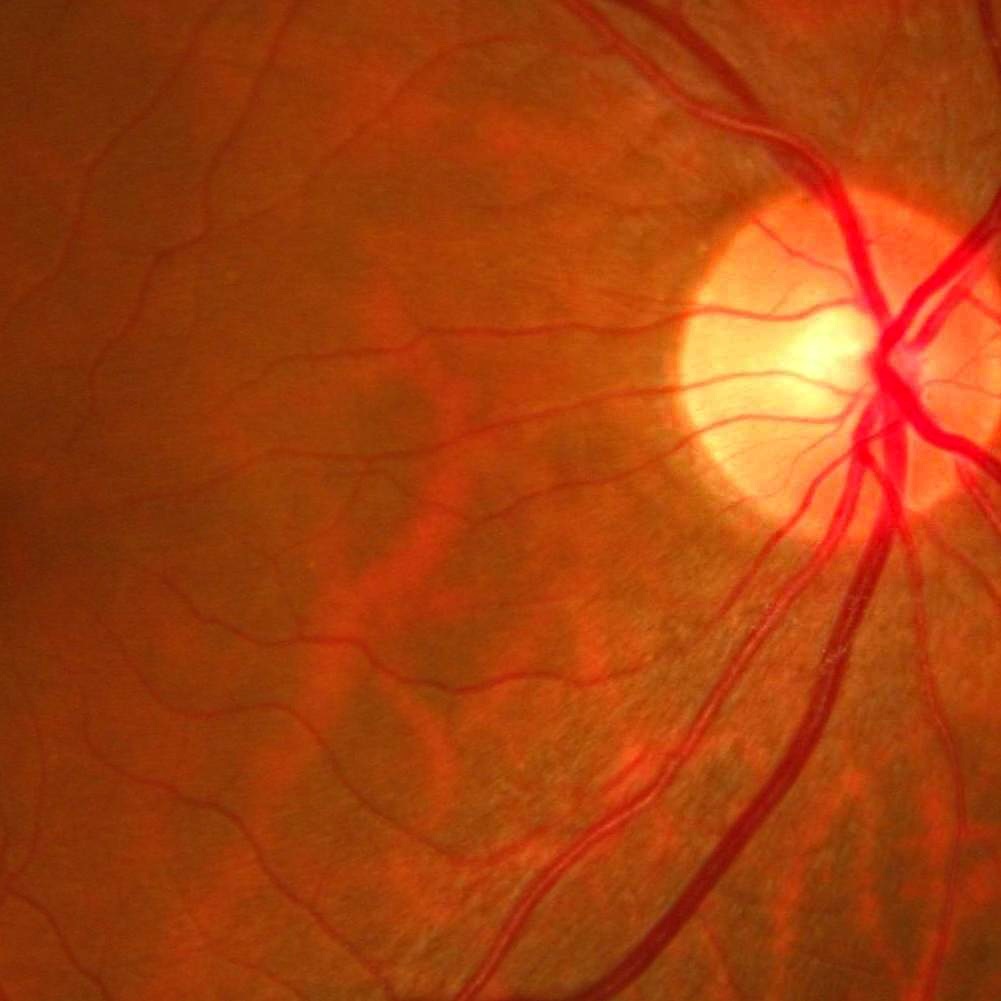}
	}} 
	\caption{  Effect of the global edge-preserving smoothing filter.
		(a) Original image (b) Output image $O^*$ without edge-preserving smoothing filter  (c) Output image after edge-preserving smoothing filter} \label{figneed}
\end{figure*}

It can be seen that  a pixel from a high variance area  will retain its values while a  pixel  from a flat area  will be smoothed by its nearby pixels. Therefore, the above averaging  often smoothes away some fine structure in regions close to flat. This is not optimal for retinal image processing and analysis. For example, the boundary between the optic cup and the neuroretinal rim is often weak. The averaging in this region might smooth away this boundary and make the optic cup segmentation more challenging. 

 To address this problem, we propose a novel structure-preserving guided retinal image filter (SGRIF)
to remove the clouding effect caused by light scattered by human-lens. 
Inspired by the GIF \cite{Hekaiming2013}, we aim to   transfer the
structure of the guidance image to the input image to preserve the edges and smooth the transferred image. 
 The proposed SGRIF is composed of a global structure transfer filter to transfer the   structure  in the retinal image and a global edge-preserving smoothing filter to smooth the transferred retinal image.   
 In our method, we use a guidance vector field $V = (V^h, V^v)$. The inputs of the proposed SGRIF are a  retinal image to be filtered and a guidance vector field. A term between the gradients of the output image $O$ and the guidance vector field $V$ is defined to penalize the original term in GIF, i.e.,
 \begin{equation}
\sum_i\|\nabla O_i -V_i\|^2  \label{eq8},
\end{equation}
where $\nabla$ denotes the operator to compute the gradients.
Combining this with first item in (\ref{eq4}), we have the objective function of the proposed global structure transfer filter:
\begin{equation}
     \lambda \sum_i(O_i-I_i)^2
+\| \nabla O_i-V_i\|^2, \label{eq9}
\end{equation} 
  where   $\lambda$ is a parameter controlling the trade-off between the two terms.
    Omitting the  subscript, the above cost function can be rewritten as:  
\begin{align}
&\lambda(O-I)^T(O-I)+(D_xO-V^h)^T(D_xO-V^h)  \nonumber \\ 
& +(D_yO-V^v)^T(D_yO-V^v),   \label{eq10}
\end{align}
where $D_x$ and $D_y$ denote discrete differentiation operators.
The output $O$ is then obtained by  solving the following linear equation:
\begin{equation}
   (\lambda A + D_x^TD_x+D_y^TD_y)O=\lambda I+D_x^TV^h+D_y^TV^v \label{eq11},
\end{equation}
where $A$ is an identity matrix.

 We solve (\ref{eq11}) using the fast separating method in \cite{Min14}. 
 The output image $O^*$ based on (\ref{eq11}) sometimes needs to be smoothed.   Fig. \ref{figneed} shows an example. As we can see,  the recovered image $O^*$ has some visual artifacts if it is not smoothed.
  To achieve  this, the output image is decomposed into two layers via an edge-preserving smoothing filter. The filter \cite{6236165, Farbman:2008:EDM:1360612.1360666, li2018} is formulated as
 \begin{equation}
   \min_{\phi}\sum_i[(\phi_i-O^*_i)^2+\gamma(\frac{(\frac{\partial\phi_i}{\partial x})^2}{|V^h_i|^{\theta}+\epsilon}+\frac{(\frac{\partial \phi_i}{\partial y})^2}{|V^v_i|^{\theta}+\epsilon})],   \label{eq12}
 \end{equation}
 where $\gamma, \theta, \epsilon$ are empirically set as 2048, 13/8 and 1/64 in all  experiments in this paper. 
   We determine the thresholds by searching from a reasonable range based on physical meaning and experience. For example $\gamma$ is set to a large value to make sure that the first term in equation  \ref{eq12} will not dominate the results. Then we conducted tests using some  $\gamma$ values and determine its value. Our results show that the small change of the parameters do not affect much the results. 
  
 We can rewrite (\ref{eq12}) as
 \begin{equation}
    (\phi-O^*)^T(\phi-O^*)+\gamma(\phi^TD_x^TB_xD_x\phi+\phi^TD_y^TB_yD_y\phi),
     \label{eq13}
 \end{equation}
 where 
 $B_x=diag\{\frac{1}{|V^h_i|^{\theta}+\epsilon}\}$,  $B_y=diag\{\frac{1}{|V^v_i|^{\theta}+\epsilon}\}$.
   
   Setting the derivative of (\ref{eq13}) to be zero, the vector $\phi$, that minimizing the above cost function, is computed by solving the following   equation:
   \begin{equation}
     (A+\gamma (D_x^TB_xD_x+D_y^TB_yD_y))\phi=O^*.    \label{eq14}
   \end{equation}
   
   Similar to that in (\ref{eq11}), the problem in (\ref{eq14}) can be solved by the  fast separate method in \cite{Min14} as well. 
   
 To apply the above models to retinal images, we need to estimate the lens-light $L_c, c\in\{r,g,b\}$. 
In this paper, we estimate $L_r$, $L_g$, and $L_b$ using the method in \cite{KIM2013410}. 
The proposed method uses the idea
of minimal color channel   and simplified dark
channel. The simplified dark channel is decomposed into
a base layer and a detail layer via the proposed method, and
the former is then used to determine the transmission map.
 The simplified dark channels of the normalized  degraded  and ideal images are computed as 
$I_c/L_c$ and $D_c/L_c$. 
Define $\tilde{I}_{min}(p)$ and $\tilde{D}_{min}(p)$ as  
\begin{equation}
   \tilde{I}_{min}(p)=\min\{\frac{I_r(p)}{L_r},\frac{I_g(p)}{L_g},\frac{I_b(p)}{L_b} \},  \label{eq15}
\end{equation}

   \begin{equation}
   \tilde{D}_{min}(p)=\min\{\frac{D_r(p)}{L_r},\frac{D_g(p)}{L_g},\frac{D_b(p)}{L_b} \}.  \label{eq16}
   \end{equation}
    It is noted that we do not consider the difference among the RGB channels in this paper, though some earlier work \cite{Marrugo11} shows that blue channel  may have more noise.   
   Since the transmission map $t$ is independent to the color channels, we have
   \begin{equation}
     \tilde{I}_{min}(p)=(1-t(p))+\tilde{D}_{min}(p)t(p).  \label{eq17}
   \end{equation}

   Let $W_{\kappa}(p)$ be a $\kappa\times \kappa$  window centered at pixel $p$.
   The  simplified dark channels of the normalized images are computed as: 
   \begin{equation}
      J_d^{\tilde{D}}(p)=\min_{p'\in W_{\kappa}(p)} \{\tilde{D}_{min}(p') \},  \label{eq18}
   \end{equation}
   
   \begin{equation}
   J_d^{\tilde{I}}(p)=\min_{p'\in W_{\kappa}(p)} \{\tilde{I}_{min}(p') \}.  \label{eq19}
   \end{equation}
   Since $t(p)$ is usually consistent within $W(p)$, 
   we have 
   
     \begin{equation}
   J_d^{\tilde{I}}(p)= (1-t(p))+J_d^{\tilde{D}}(p)  t(p). \label{eq20}
   \end{equation}
   
\begin{figure} 
 	\centering
 	{\subfigure[]{
 			\includegraphics[height=1.6in, width=1.6in]{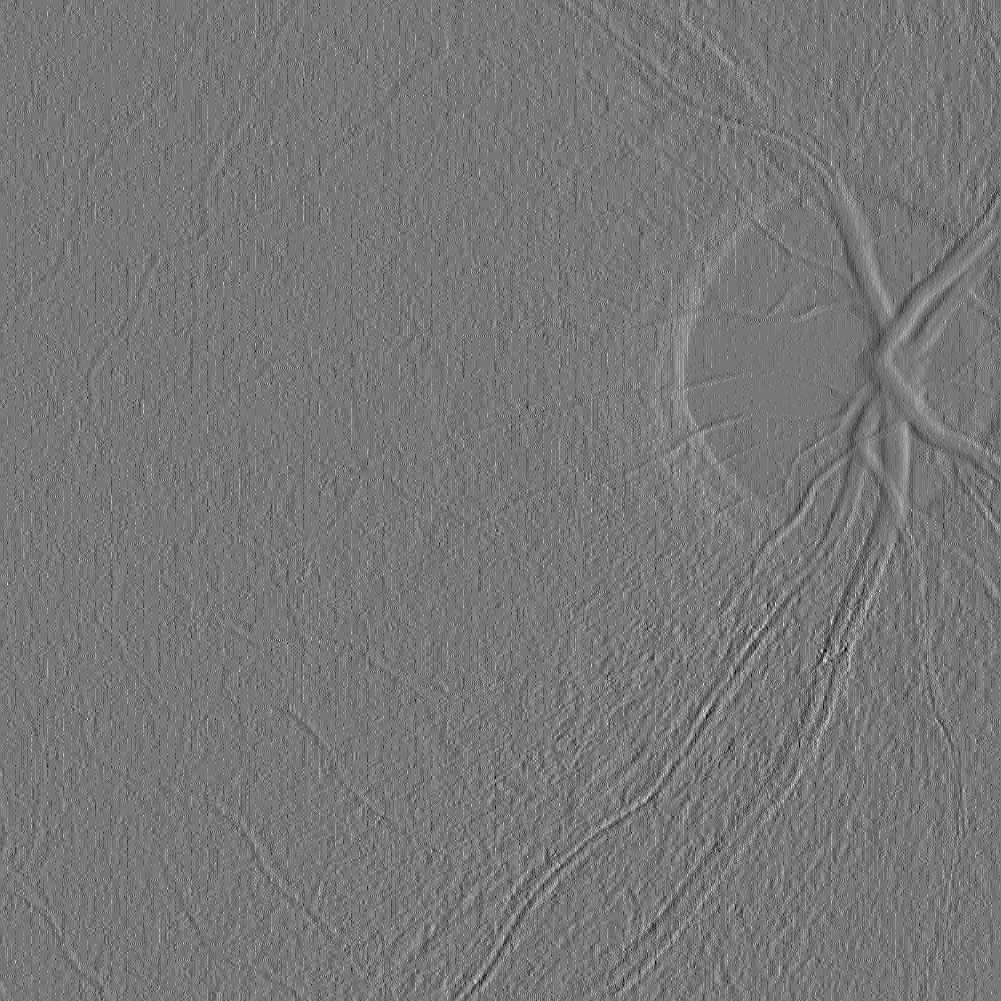}
 	}} 
 	{\subfigure[]{
 			\includegraphics[height=1.6in, width=1.6in]{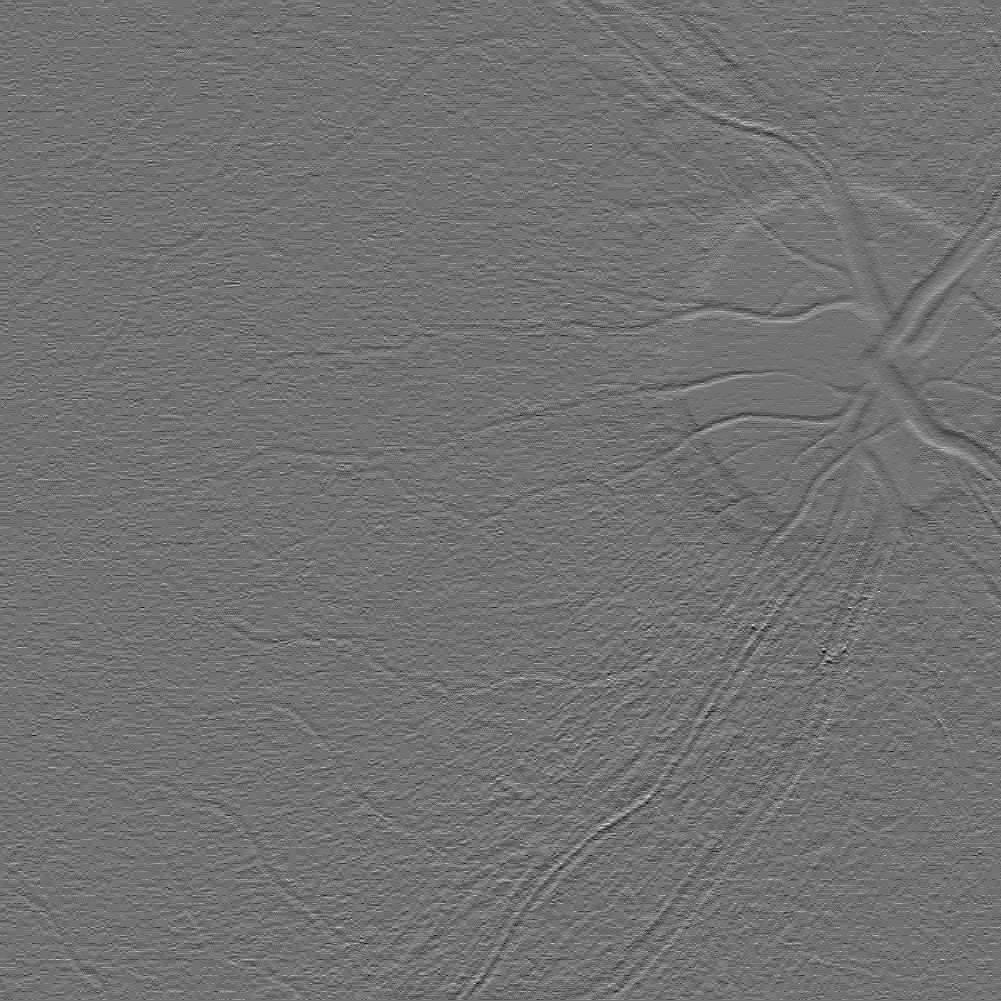}
 	}} 
 	 \caption{An example of the gradient vector field image.  (a)   magnitude of    $V^h$, (b)   magnitude of     $V^v$ }  \label{figexc}
 \end{figure}

The guidance vector field $V = (V^h, V^v)$ is computed as 
  \begin{equation}
   V^h(m,n) = \tilde{I}_{min}(m,n+1)- \tilde{I}_{min}(m,n), 
 \end{equation}
 \begin{equation}
  V^v(m,n) = \tilde{I}_{min}(m+1,n)- \tilde{I}_{min}(m,n). 
 \end{equation} 
 An example of the gradient vector field image computed from Fig. \ref{figneeda} is shown in Fig. \ref{figexc}.
   Combining the gradient vector field with equation (\ref{eq9}), we obtain the output $O^*$.

  We further smooth the first item $\phi(p)=1-t(p)$ using the edge preserving smoothing filter 
  in  (\ref{eq12}) and obtain $\phi^*(p)$. 
   
 The transmission map $t(p)$ is then computed as:
 \begin{equation}
     t^*(p)=1-\phi^*(p).  \label{eq21}
 \end{equation}
 The underlying image is computed as:
 \begin{equation}
 D_c(p)=\frac{I_c(p)-L_c}{t^*(p)}+L_c.  \label{eq22}
 \end{equation}
 
 In our model, we ignore the attenuation $\alpha$. In the cases where the attenuation $\alpha$ cannot be ignored, we can still solve the problem in the similar way except that we need to estimate $\alpha$ using a precataract image and compute the final output image as  $\frac{1}{\alpha}D_c(p)$. In this paper, we simply restore the image based on equation (\ref{eq22}).

 \section{Experimental Results} \label{results}
 \subsection{Data Set}

We conduct experiments using the 650 images from the ORIGA data set \cite{origa}\cite{Cheng:17}. In the 650 images, there are 203 images from eyes with cataracts and 447 images from eyes without any cataract.   We mainly apply SGRIF on the region around the optic disc and evaluate how it affects   subsequent analysis on optic disc.  In this paper, we use the disc from the original image and the disc boundaries are kept the same for all three images. This is to prevent the error propagation from optic disc segmentation to optic cup segmentation. 
\subsection{Evaluation Metrics}

To evaluate the performance of the proposed method, we first compute how it affects the contrast of the optic disc image. Two evaluation metrics, namely the histogram flatness measure （(HFM) and the histogram spread (HS) have been proposed to evaluate the contrast of images previously \cite{Tripathi11}:
  \begin{equation}
HFM=\frac{(\prod_{i=1}^nx_i)^{\frac{1}{n}}}{\frac{1}{n}\sum_{i=1}^nx_i},
\end{equation}
 where $x_i$ is the histogram count for the $i^{th}$ histogram bin
and $n$ is the total number of histogram bins. 
\begin{equation}
    HS=\frac{(3^{rd} \text{quartile}-1^{st} \text{quartile}) \text{ of histogram}}{\text{(maximum-minimum) of the pixel value range}}.
\end{equation}

We also compute the mean variability of the local luminosity (VLL) \cite{FORACCHIA2005179} throughout the optic disc.
Given an image $I$, we divide it into $N\times N$ blocks $B_{i,j},
i,j = 1, \cdots, N$ with equal sizes. We first compute the mean of block $B_{i,j}$ as $\mu(i,j)$.
%

VLL is
computed as

\begin{equation}
    VLL = \frac{1}{N}\sqrt{\frac{1}{\bar{I}^2}\sum_{i=1}^N\sum_{j=1}^N (\mu(i,j)-\bar{I})^2},
\end{equation}
where $\bar{I}$ stands for the mean intensity of the entire image.  

 For all the above metrics, a high value indicates a better result.

\begin{figure*}
	\centering
 	{\subfigure{
			\includegraphics[height=1.3in, width=1.3in]{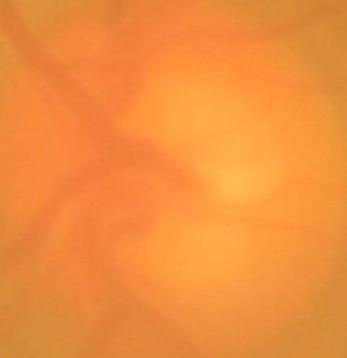}
	}}\hspace{-0.25cm}
	{\subfigure{
			\includegraphics[height=1.3in, width=1.3in]{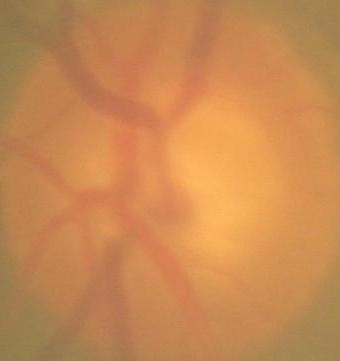}
	}}\hspace{-0.25cm}
	{\subfigure{
			\includegraphics[height=1.3in, width=1.3in]{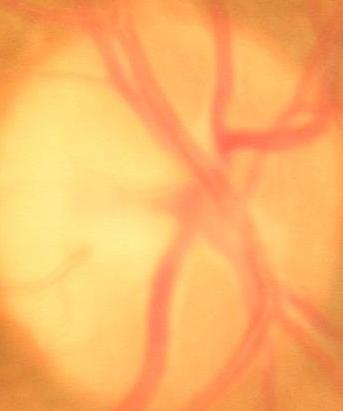}
	}}\hspace{-0.25cm}
	 	{\subfigure{
	 		\includegraphics[height=1.3in, width=1.3in]{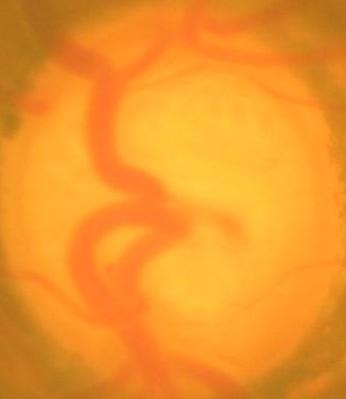}
	 }}\hspace{-0.25cm}
  		{\subfigure{
  			\includegraphics[height=1.3in, width=1.3in]{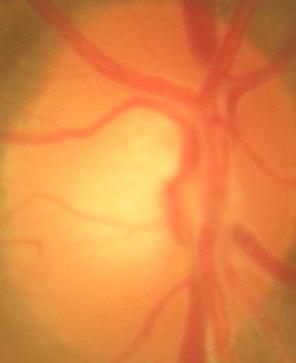}
  	}}\\
 	{\subfigure{
			\includegraphics[height=1.3in, width=1.3in]{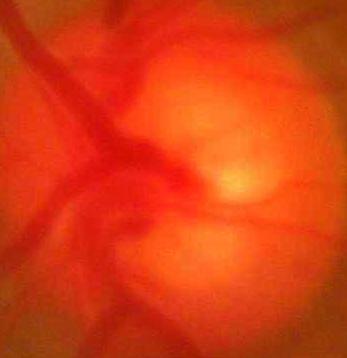}
	}}\hspace{-0.25cm}
 {\subfigure{
		\includegraphics[height=1.3in, width=1.3in]{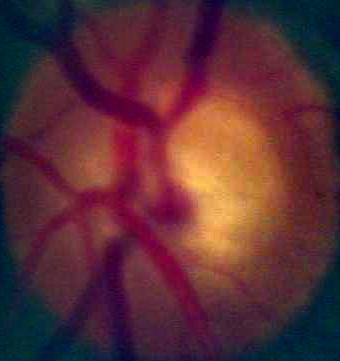}
}}\hspace{-0.25cm}
	{\subfigure{
		\includegraphics[height=1.3in, width=1.3in]{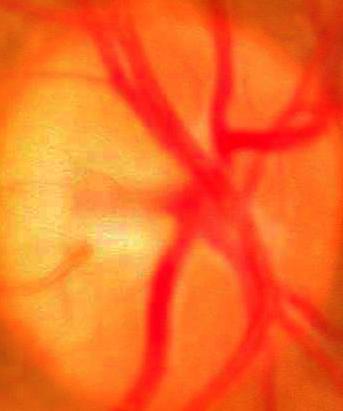}
}}\hspace{-0.25cm}
	{\subfigure{
		\includegraphics[height=1.3in, width=1.3in]{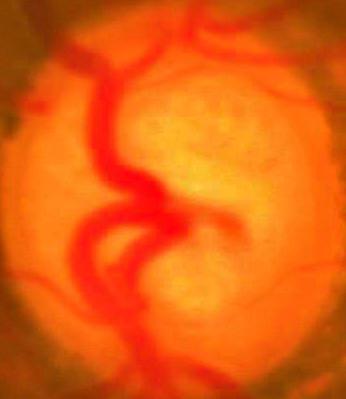}
}}\hspace{-0.25cm}
	{\subfigure{
		\includegraphics[height=1.3in, width=1.3in]{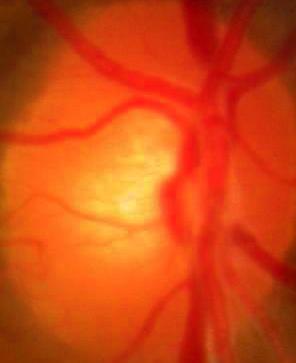}
}}\\
{\subfigure{
		\includegraphics[height=1.3in, width=1.3in]{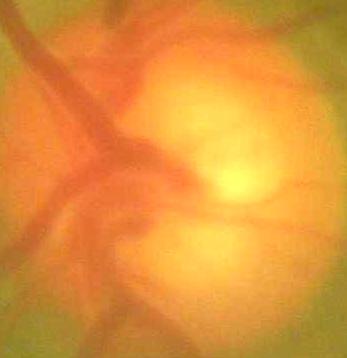}
}}\hspace{-0.25cm}
{\subfigure{
		\includegraphics[height=1.3in, width=1.3in]{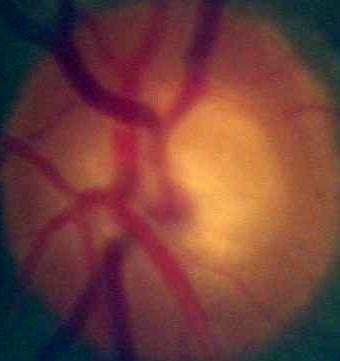}
}}\hspace{-0.25cm}
{\subfigure{
		\includegraphics[height=1.3in, width=1.3in]{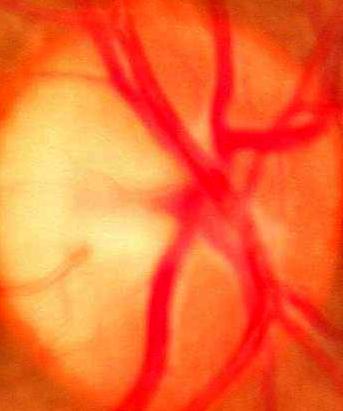}
}}\hspace{-0.25cm}
	{\subfigure{
		\includegraphics[height=1.3in, width=1.3in]{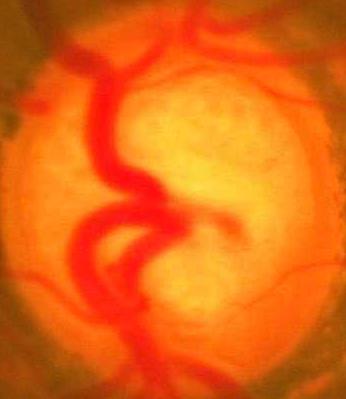}
}}\hspace{-0.25cm}
{\subfigure{
		\includegraphics[height=1.3in, width=1.3in]{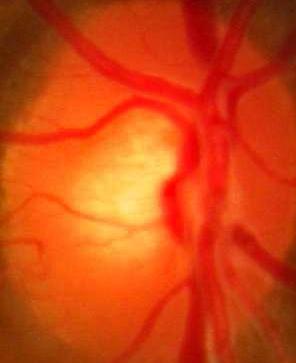}
}}\\
	\caption{Retinal images of optic disc: first row the original disc images; second row the images processed by GIF and last row the images processed by proposed filter.} \label{figdisc}
\end{figure*}

\begin{table*}
	\caption{ Performance by various methods.
	} \begin{center}\
		\begin{tabular}{c|c|c| c|c|c|c |c|c|c  } \hline
			& \multicolumn{3}{|c}{Histogram Flatness Measure }    &  \multicolumn{3}{|c}{ Histogram Spread} &\multicolumn{3}{|c}{ Variability of Local Luminosity}
			\\\hline
			& All &  Cataract   & No Cataract & All &  Cataract & No Cataract & All &  Cataract   & No  Cataract   \\\hline
			Original   &  0.5786 & 0.5295 &  0.6009 &   0.2856 & 0.2692  & 0.2930 & 0.0517 & 0.0363 & 0.0537\\\hline
		     GIF \cite{Hekaiming2013}   & 0.4779 & 0.4324 & 0.4986  &  0.2505 & 0.2329 & 0.2584 & 0.1014 &  0.0999 & 0.1021 \\
			\hline
			\textbf{Proposed } & \textbf{0.6129}  &    \textbf{0.5817} &  \textbf{0.6271}     &   \textbf{0.2980} & \textbf{0.2806}   & \textbf{0.3059} &\textbf{0.1214}  &  \textbf{0.1076} & \textbf{0.1277} \\\hline
		\end{tabular}
	\end{center}
	\label{table1}
\end{table*}
\subsection{Results}

Table  \ref{table1}  summarizes the results, from which we can see that the proposed method improves the HFM, HS and VLL by 5.9\%, 4.3\% and 134.8\%, respectively compared with original images.  GIF improves VLL, but it does not increase HFM and HS. This is because GIF oversmoothes some region close to flat and reduces the dynamic range of the histograms. 
   Fig. \ref{figdisc} shows results from five sample images of optic disc.
   As we can see, the proposed SGRIF enhances the contrast between the optic cup and the neuroretinal rim while the improvement by GIF is less clear.
    Visually, it is difficult to tell if GIF has oversmoothed some regions but we will show the differences from subsequent  analysis below.

\subsection{Application}
In order to show the benefits of the declouding, we further conduct experiments to examine how the declouding processing affects the analysing tasks. Two slightly different applications are used as examples: 1) deep learning based optic cup segmentation; 2) sparse learning based cup to disc ratio measurement.

\subsubsection{Deep Learning based Optic Cup Segmentation}
\begin{figure}
		\includegraphics[height=1.45in]{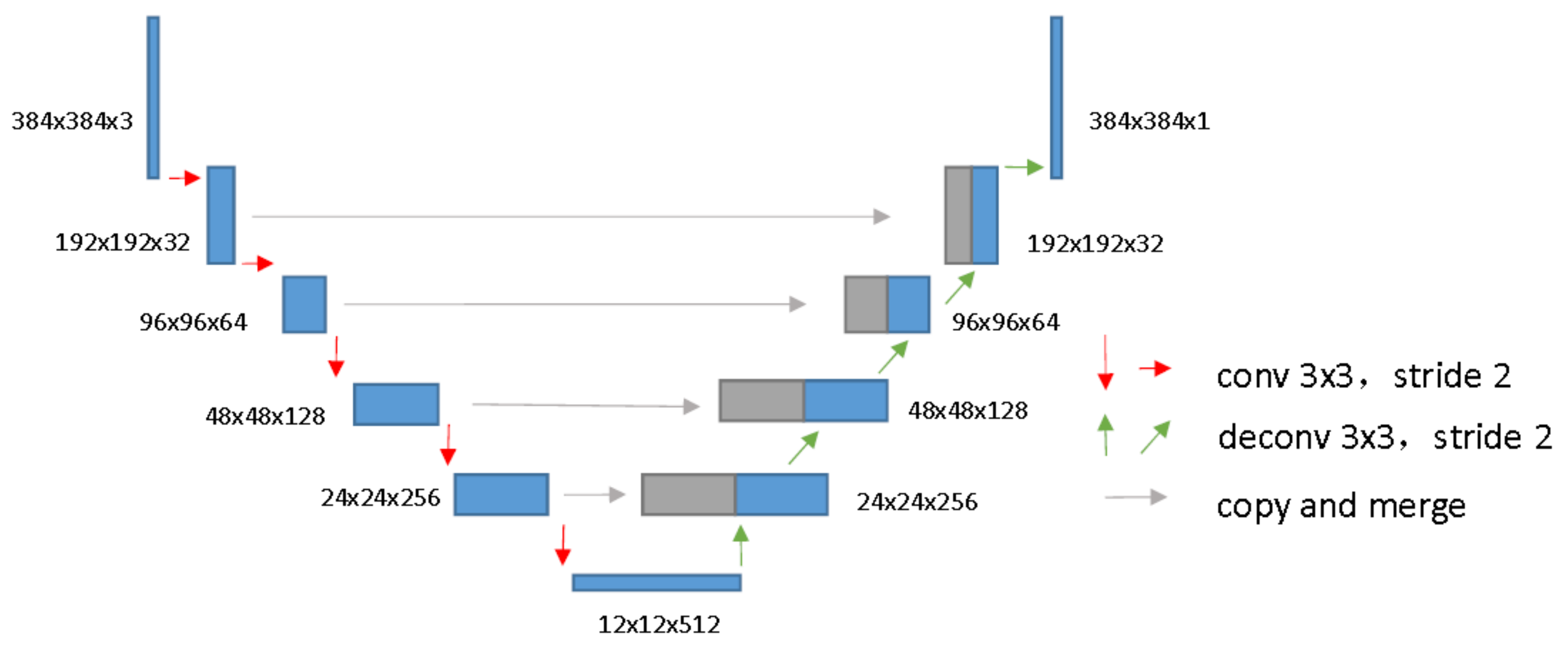} 
			\caption{Simple U-Net architecture: }\label{figunet}
\end{figure}

In the first example, we examine how the decloud affects the optic cup segmentation. Since deep learning \cite{Lecun15} has shown to be promising in segmentation, we use deep learning as the baseline approach and the U-Net architecture \cite{Ronneberger2015} is adopted in this paper. U-Net is a fully convolution neural network for the biomedical image segmentation. In our implementation, we use a simplified U-Net as the original U-net requires much more number of parameters to be trained.   Our network is illustrated in Fig. \ref{figunet}. Similar to original U-Net, our network contains an encoder and a decoder. For each layer of the encoder, it just adopts a convolution layer with stride 2, replacing the original two convolution layers and one pooling layer. For each layer of the decoder, it takes two outputs as its input: i) the output of last layer in the encoder; ii) the corresponding layer in the encoder with the same size. Then two middle-layer feature maps are concatenated and transferred to the next deconvolution layer as its input. Finally, our simplified U-Net outputs a prediction grayscale map which ranges from 0 to 255. We calculate the mean square error between prediction map and ground truth  as loss function. 
In simple U-Net, we use a mini-batch Stochastic Gradient Decent (SGD) algorithm to optimize the loss function, specifically, Adagrad based SGD \cite{Duchi:2011:ASM:1953048.2021068}. The adopted learning rate is 0.001 and batch size is 32 under the Tensorflow framework based on Ubuntu 16.04 system.   We
use a momentum of 0.9. All images are resized to $384\times 384$ for training and testing.  The obtained probability map is resized back to original  size to obtain the segmentation results. 

In our experiments, the 650 images have been
divided randomly into set A of 325  training  images  and set
B of 325 testing images. In order to
evaluate the performance of the proposed method for images from eyes with
and without cataract, the 325 images in set B is divided into  a subset of 113 images with cataracts and 212 images without any cataract.  We then compare the output cup segmentation results in the three different scenarios: 1) Original: the original training and testing images are used for training and testing the U-Net model; 2) GIF: all the training and testing images are first filtered by GIF before U-Net model training and testing 3) SGRIF: all the training and testing images are first filtered by SGRIF before U-Net model training and testing. 

\begin{figure*}
	\centering
 	{\subfigure{
			\includegraphics[height=1.3in, width=1.3in]{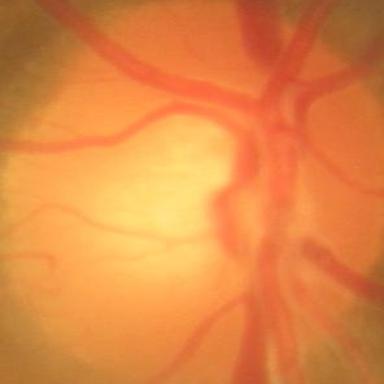}
	}}\hspace{-0.25cm}
	{\subfigure{
			\includegraphics[height=1.3in, width=1.3in]{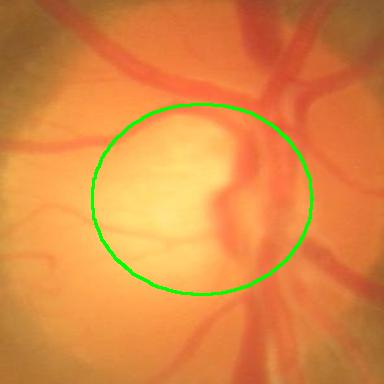}
	}}\hspace{-0.25cm}
	{\subfigure{
			\includegraphics[height=1.3in, width=1.3in]{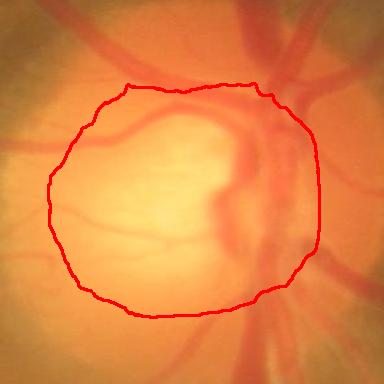}
	}}\hspace{-0.25cm}
	{\subfigure{
			\includegraphics[height=1.3in, width=1.3in]{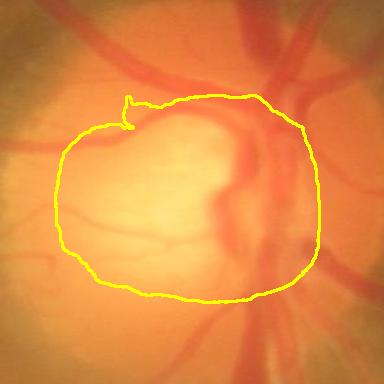}
	}}\hspace{-0.25cm}
	{\subfigure{
			\includegraphics[height=1.3in, width=1.3in]{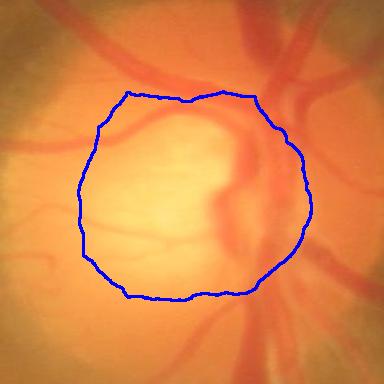}
	}}
	{\subfigure{
		\includegraphics[height=1.3in, width=1.3in]{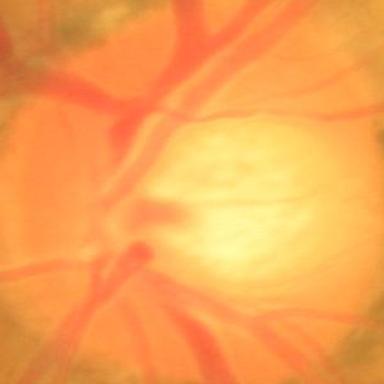}
}}\hspace{-0.25cm}
{\subfigure{
		\includegraphics[height=1.3in, width=1.3in]{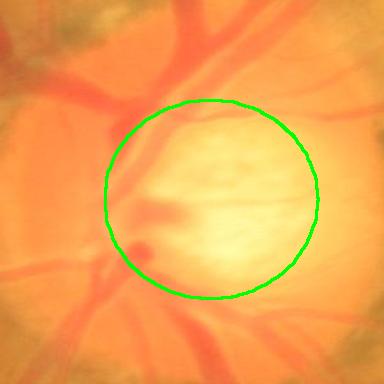}
}}\hspace{-0.25cm}
{\subfigure{
		\includegraphics[height=1.3in, width=1.3in]{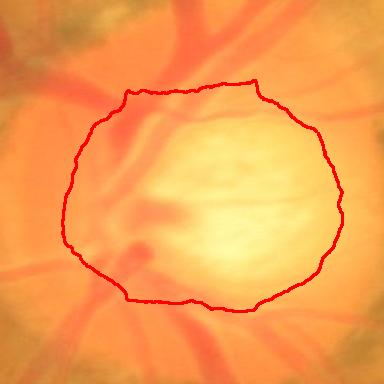}
}}\hspace{-0.25cm}
{\subfigure{
		\includegraphics[height=1.3in, width=1.3in]{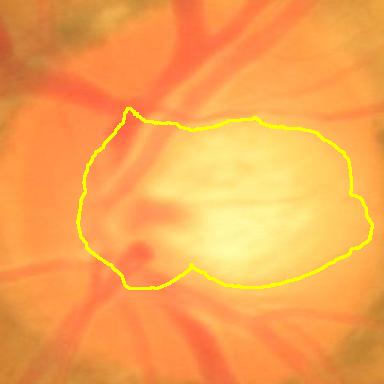}
}}\hspace{-0.25cm}
{\subfigure{
		\includegraphics[height=1.3in, width=1.3in]{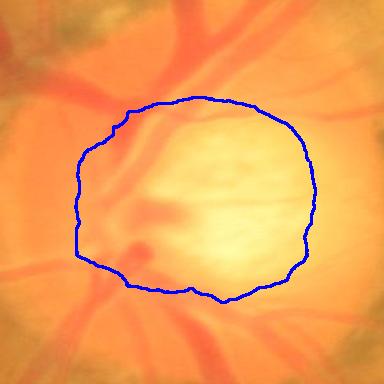}
}}
{\subfigure{
		\includegraphics[height=1.3in, width=1.3in]{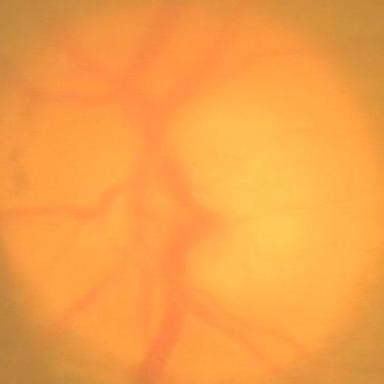}
}}\hspace{-0.25cm}
{\subfigure{
		\includegraphics[height=1.3in, width=1.3in]{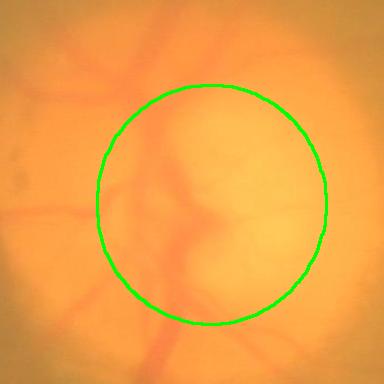}
}}\hspace{-0.25cm}
{\subfigure{
		\includegraphics[height=1.3in, width=1.3in]{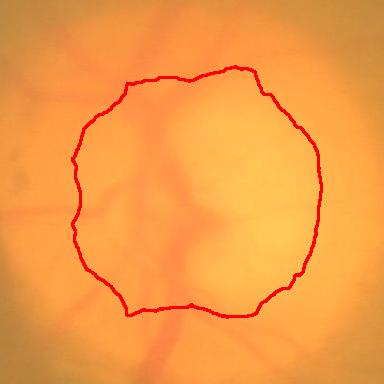}
}}\hspace{-0.25cm}
{\subfigure{
		\includegraphics[height=1.3in, width=1.3in]{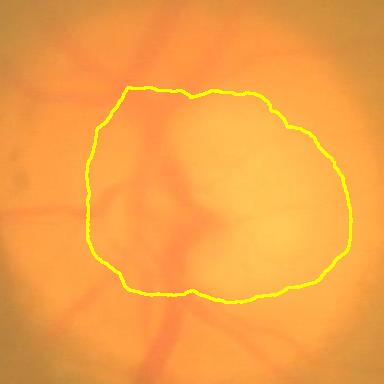}
}}\hspace{-0.25cm}
{\subfigure{
		\includegraphics[height=1.3in, width=1.3in]{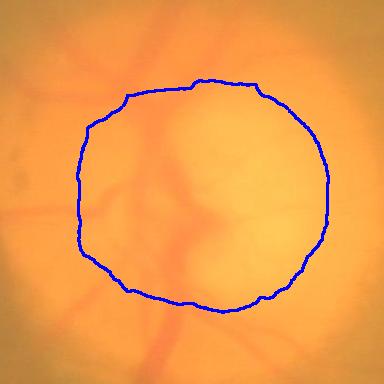}
}}
{\subfigure{
		\includegraphics[height=1.3in, width=1.3in]{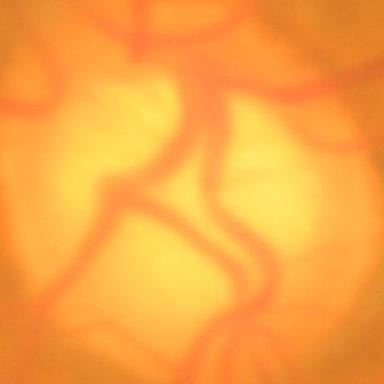}
}}\hspace{-0.25cm}
{\subfigure{
		\includegraphics[height=1.3in, width=1.3in]{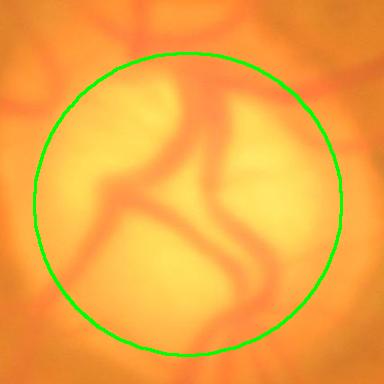}
}}\hspace{-0.25cm}
{\subfigure{
		\includegraphics[height=1.3in, width=1.3in]{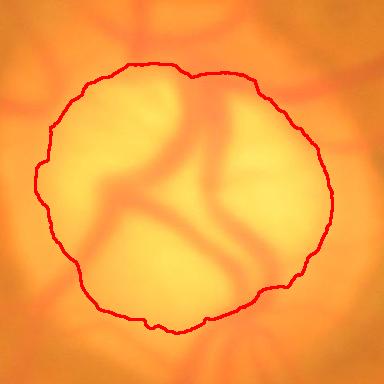}
}}\hspace{-0.25cm}
{\subfigure{
		\includegraphics[height=1.3in, width=1.3in]{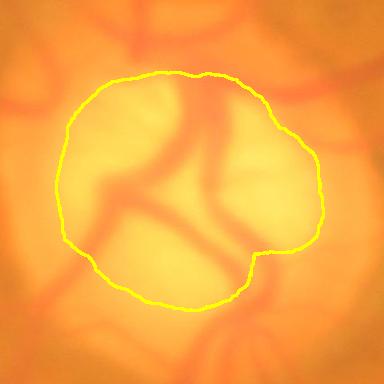}
}}\hspace{-0.25cm}
{\subfigure{
		\includegraphics[height=1.3in, width=1.3in]{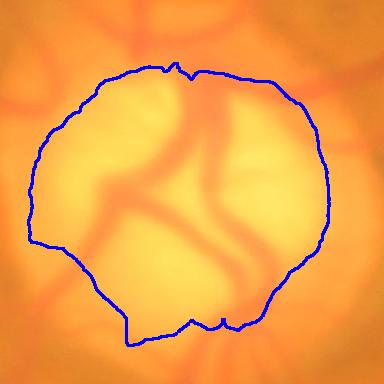}
}}\setcounter{subfigure}{0}
{\subfigure[]{
		\includegraphics[height=1.3in, width=1.3in]{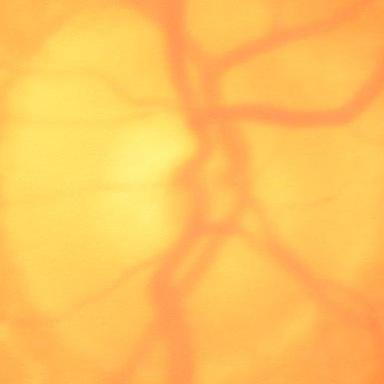}
}}\hspace{-0.25cm}
{\subfigure[]{
		\includegraphics[height=1.3in, width=1.3in]{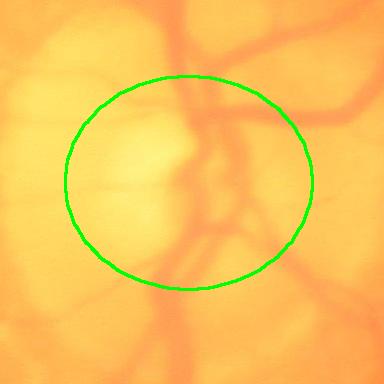}
}}\hspace{-0.25cm}
{\subfigure[]{
		\includegraphics[height=1.3in, width=1.3in]{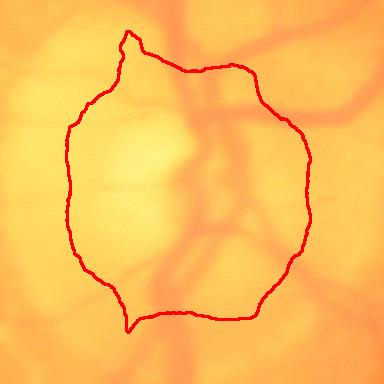}
}}\hspace{-0.25cm}
{\subfigure[]{
		\includegraphics[height=1.3in, width=1.3in]{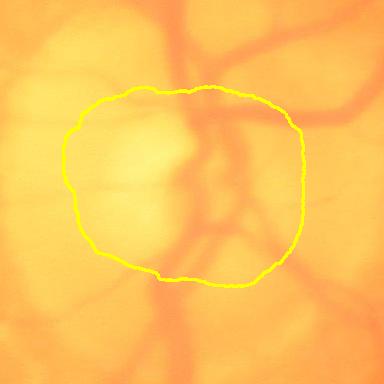}
}}\hspace{-0.25cm}
{\subfigure[]{
		\includegraphics[height=1.3in, width=1.3in]{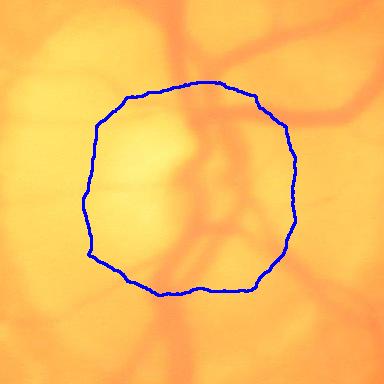}
}}\\
	\caption{Effect of  filtering on optic cup segmentation: (a) original disc (b) manual ground truth cup boundary (c) segmentation based on original disc image (d) segmentation based on GIF processed disc image (e) segmentation based on proposed SGRIF processed disc image } \label{fig4}
\end{figure*}

To compare the segmentation results, we compute the  commonly used overlapping error $E$  as the evaluation metric.
  \begin{equation}  E = 1-\frac{Area(S \cap G)}{Area(S \cup
 	G)},\end{equation}
 where $S$ and  $G$ denote  the segmented and
 the manual ``ground truth" optic cup respectively.
  Table \ref{table2} summarizes the results in the three different scenarios.  As we can see, the proposed SGRIF reduces the overlapping error by 7.9\% relatively from 25.3\% without filtering image to 23.3\% with SGRIF.  The statistical t-test indicates that the reduction is significant with $p<0.001$.    However GIF reduces the accuracy in optic cup segmentation. To visualize the segmentation results, we also draw the boundaries of segmented optic cup by the deep learning models.  Fig. \ref{fig4} shows five examples where the first two are from  eyes without cataract and the last  three are from eyes with cataract. From the results, we can see that SGRIF improves the optic cup segmentation  in both images from eyes with and without cataracts.  SGRIF reduces the segmentation error as it improves the boundary at tempral side of the disc.  GIF is not very helpful as it often smoothes away the boundaries in areaes close to flat. This often happens in  the temporal side of the retinal image where the gradient change is mild.  The above observation is inline with our discussion in Section \ref{spgrif} that GIF might smooth away the weak boundaries and make the optic cup segmentation more challenging. 
  
\begin{table}
	\caption{ Overlapping Error in Optic Cup Segmentation
	} \begin{center}\
		\begin{tabular}{c|c|c| c    } \hline
						& All &  Cataract   & No Cataract    \\\hline
			Original   &  25.3 &  24.4 &   25.8      \\\hline
			GIF \cite{Hekaiming2013}   & 25.6 & 24.8 & 26.0    \\
			\hline
 			\textbf{Proposed } & \textbf{23.3}  &    \textbf{22.8} &  \textbf{23.5}      \\\hline	
		\end{tabular}
	\end{center}
	\label{table2}
\end{table}

\begin{figure*}
	\centering
	{\subfigure{
			\includegraphics[height=2in, width=2in]{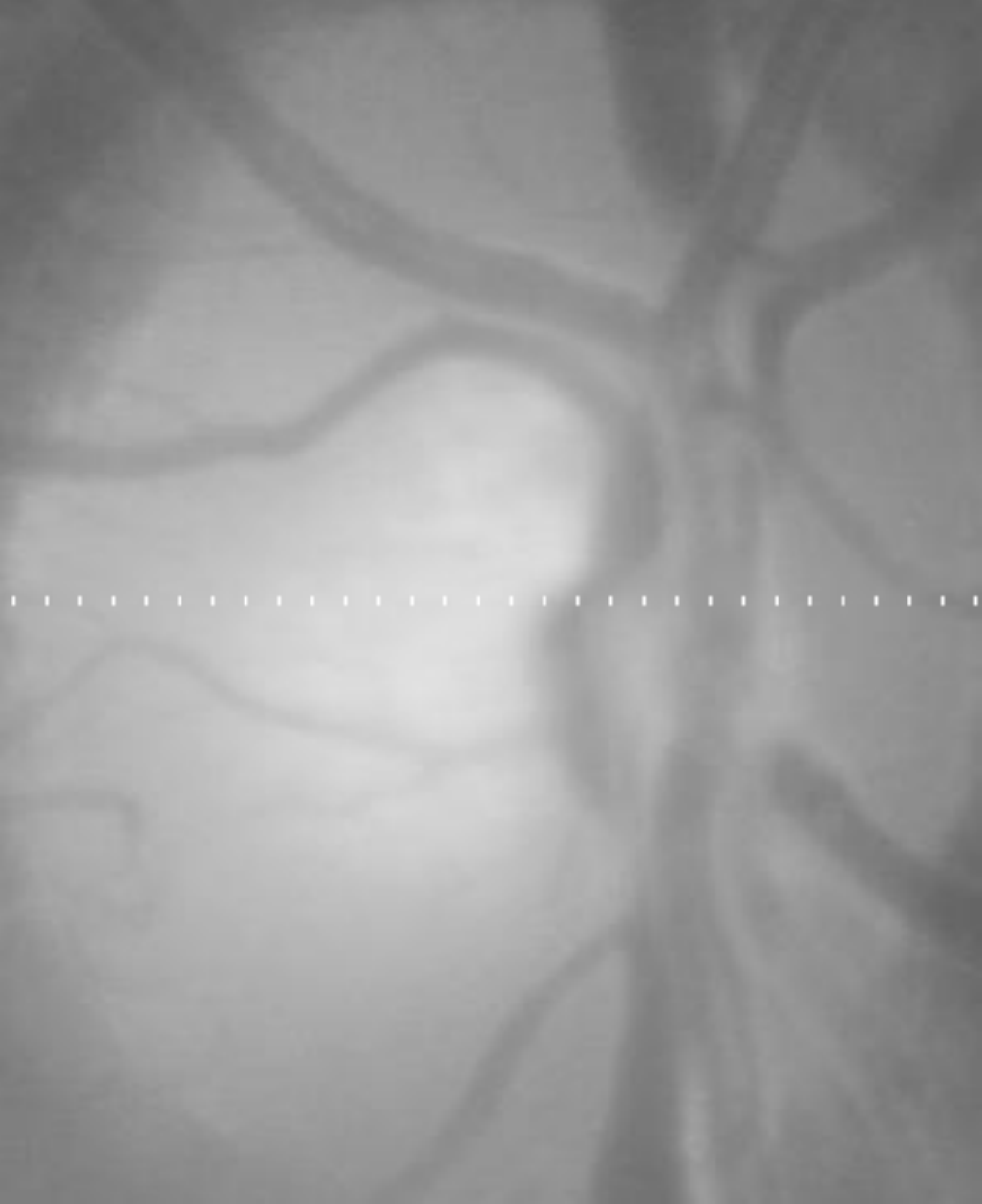}
	}}
	{\subfigure{
			\includegraphics[height=2in, width=2in]{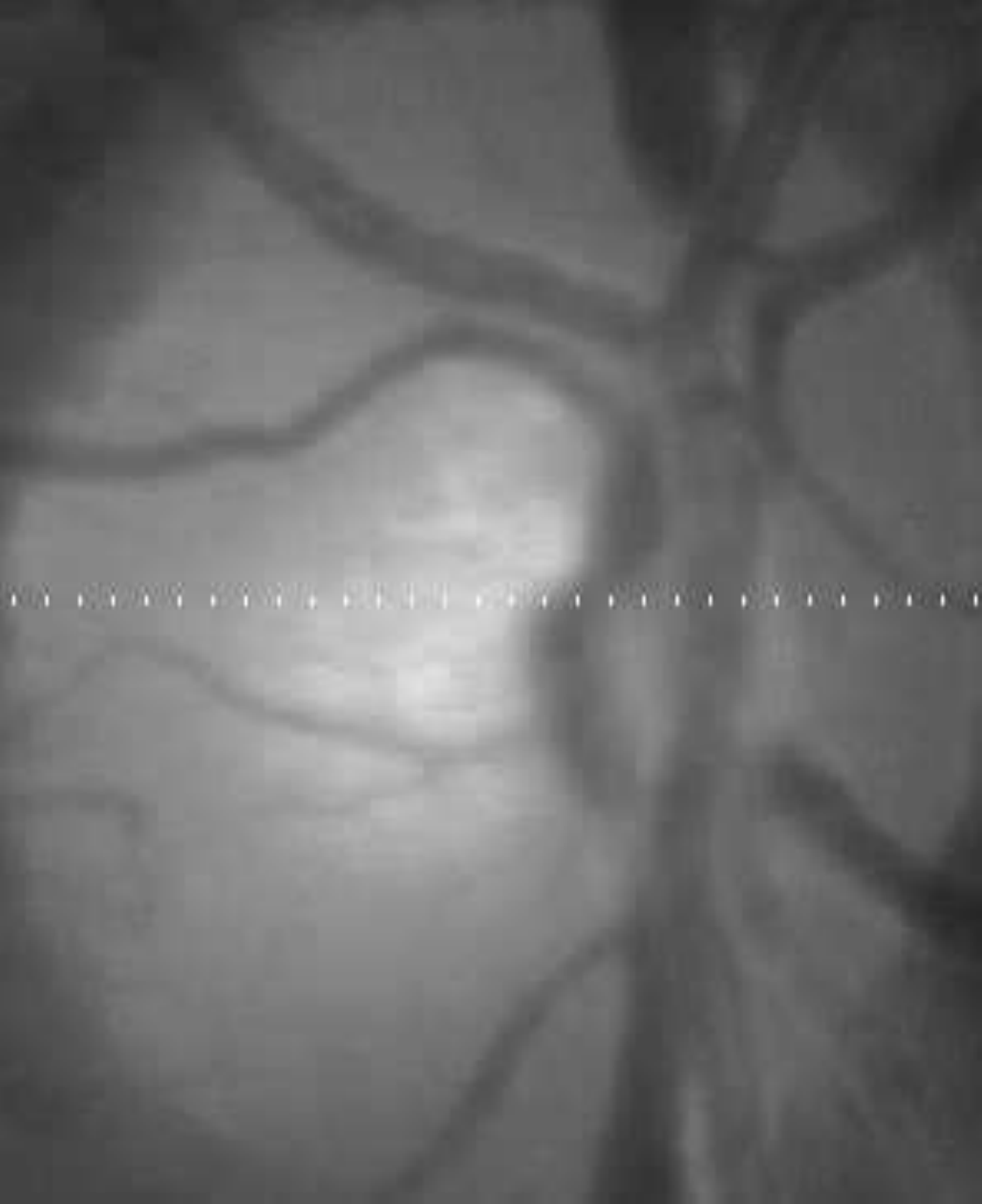}
	}}  
	{\subfigure{
			\includegraphics[height=2in, width=2in]{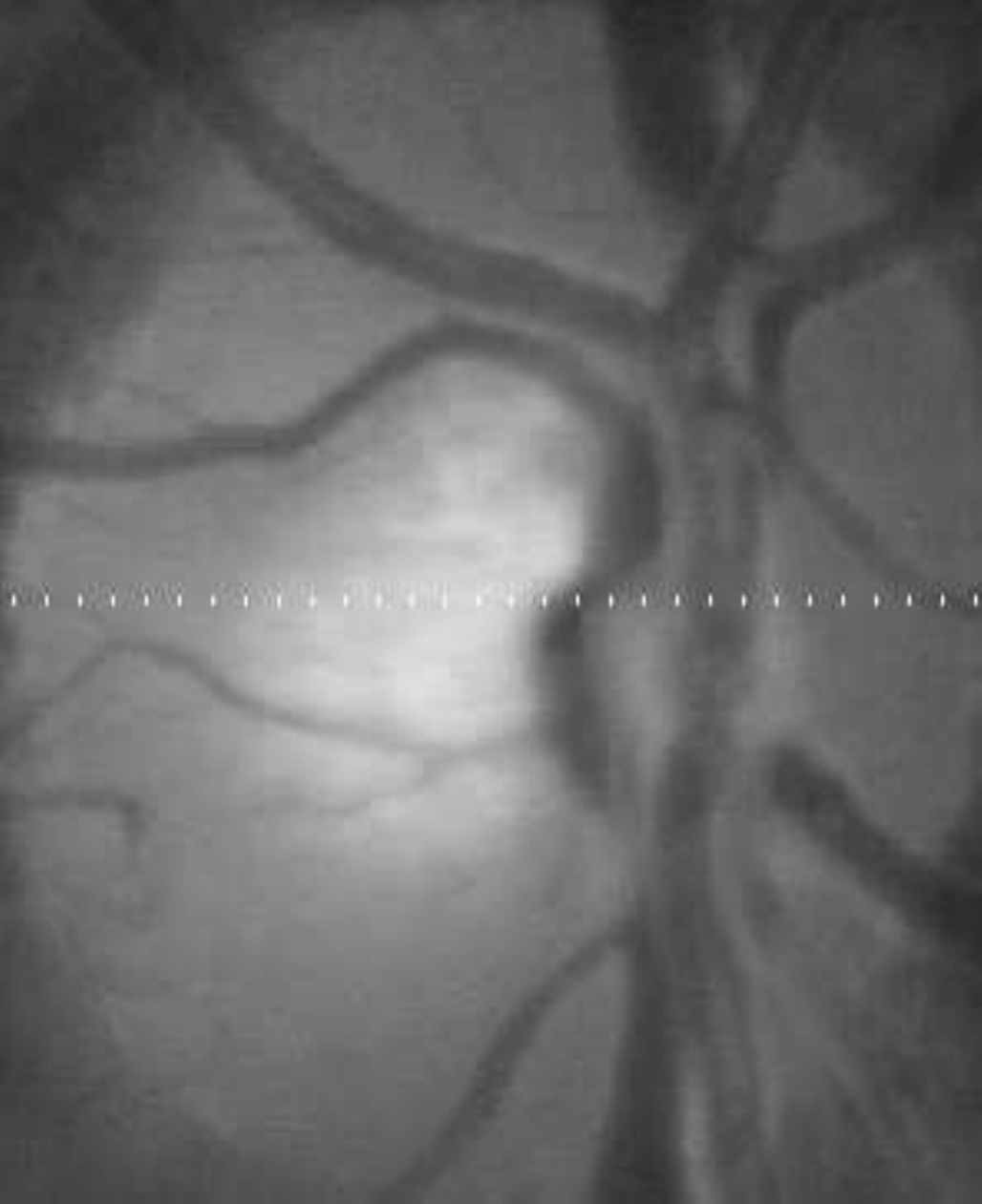}
	}} 
	{\subfigure{
		\includegraphics[height=1.5in, width=6in]{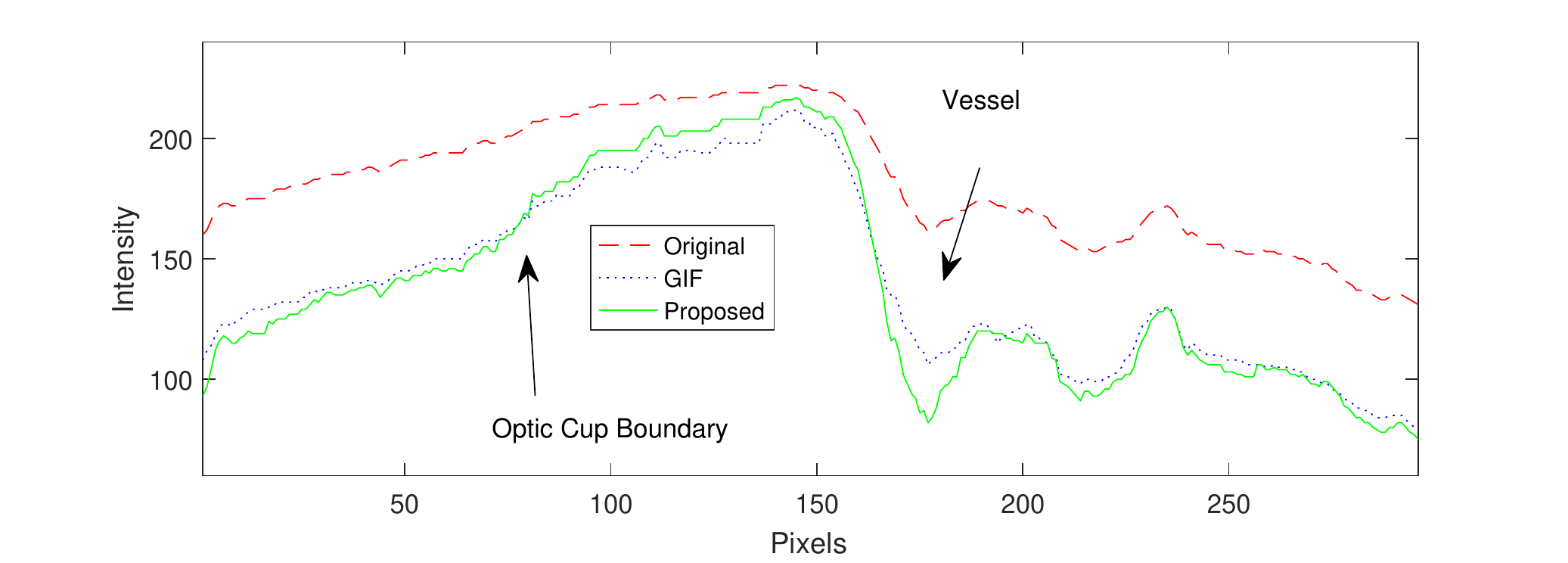}
}}
	{\subfigure{
			\includegraphics[height=2in, width=2in]{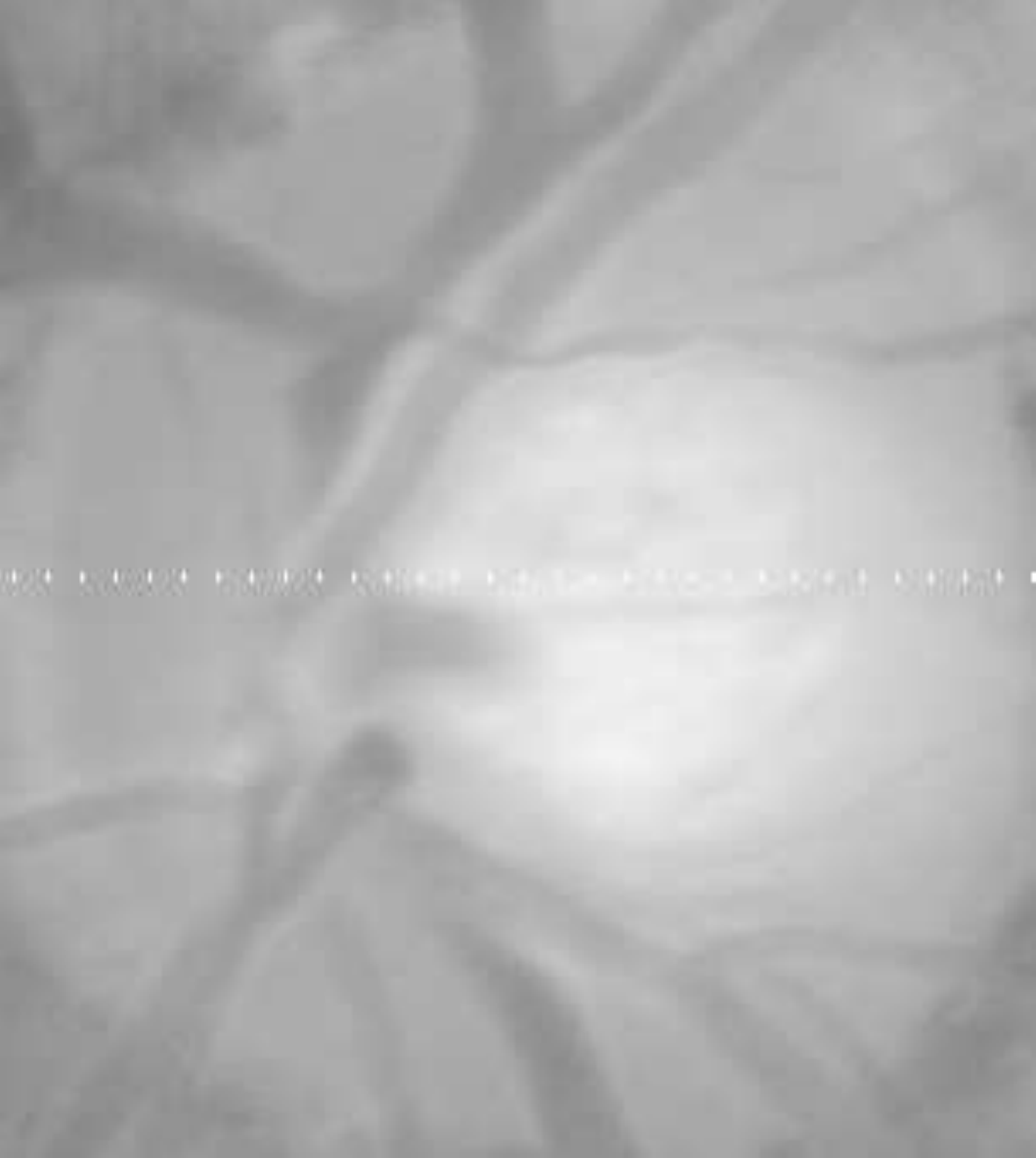}
	}}
	{\subfigure{
			\includegraphics[height=2in, width=2in]{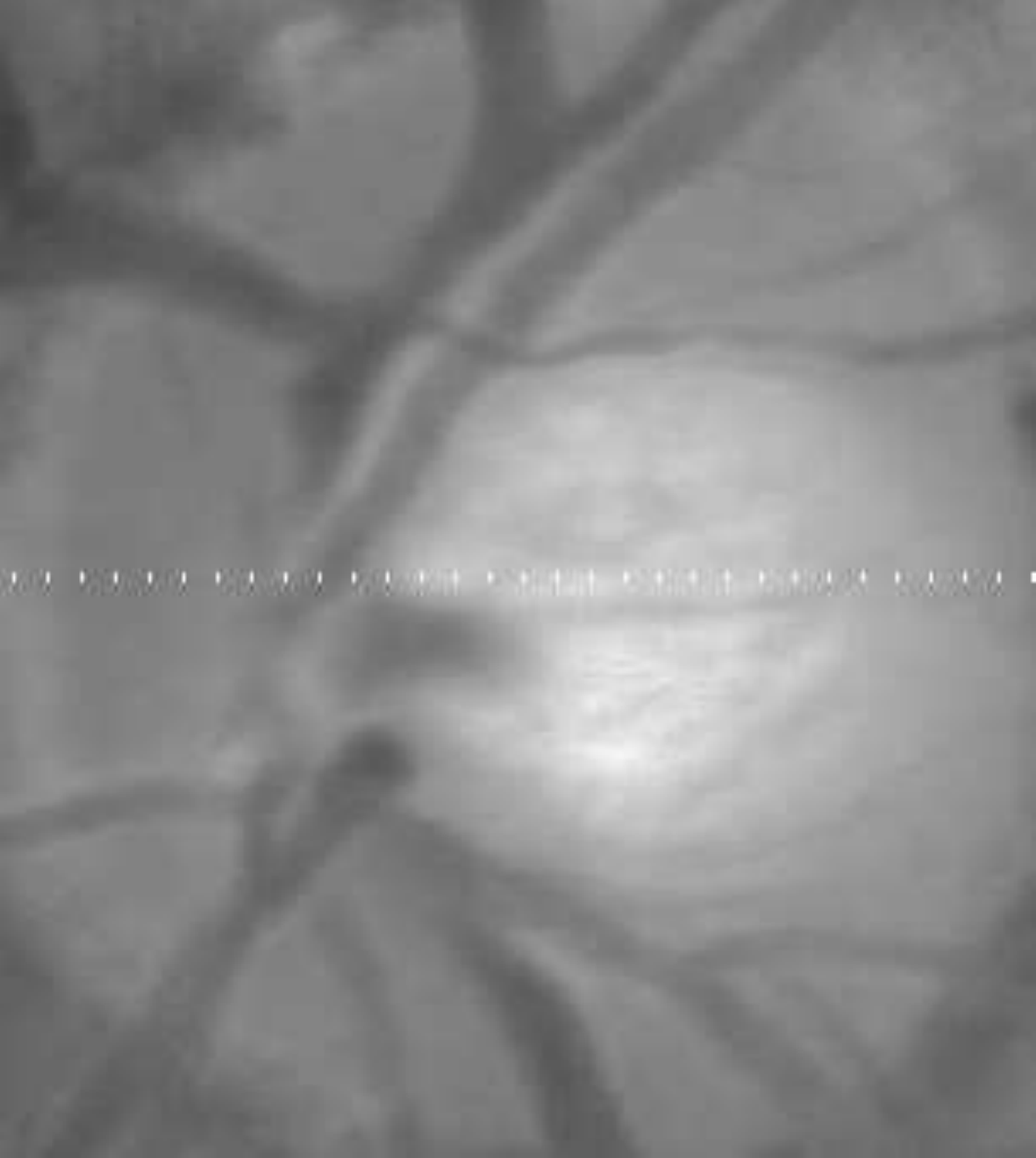}
	}}  
	{\subfigure{
			\includegraphics[height=2in, width=2in]{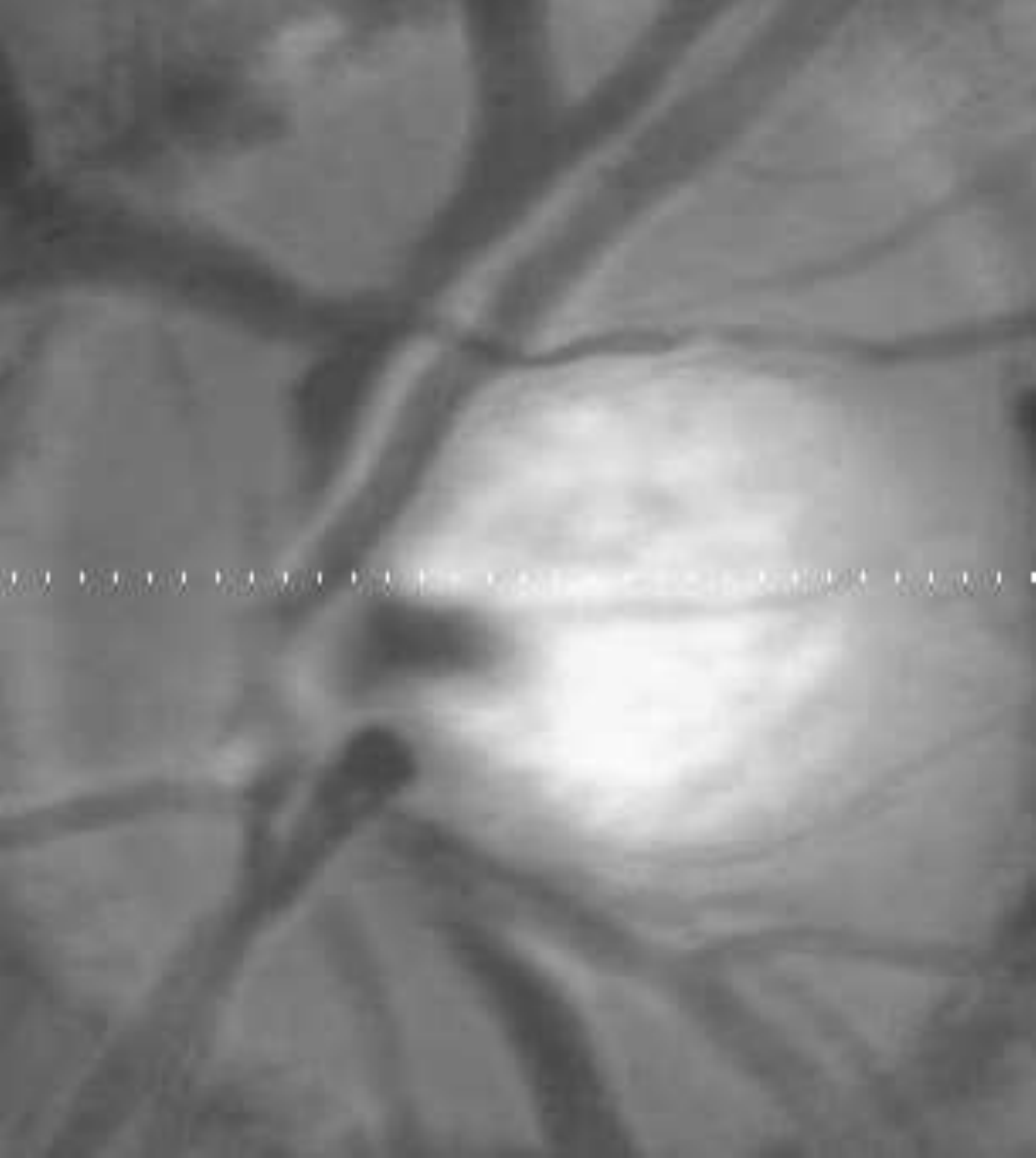}
	}}\\
	{\subfigure{
		\includegraphics[height=1.5in, width=6in]{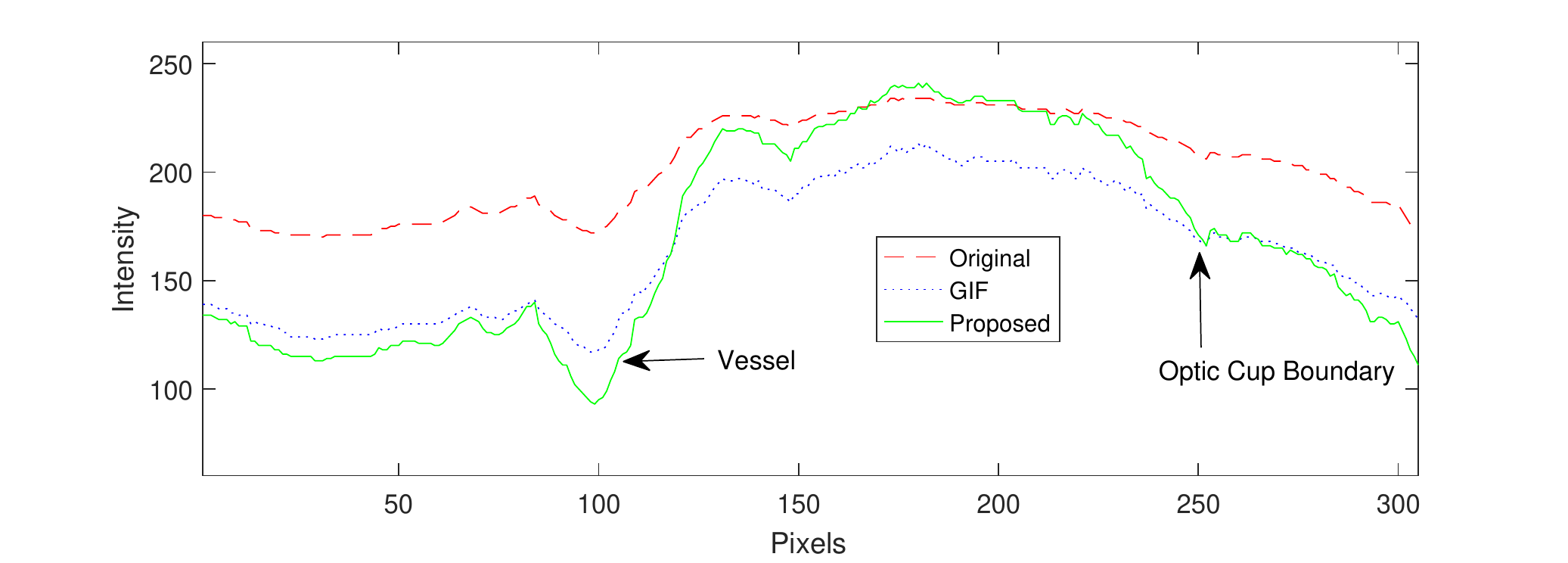}
}}
 
  \caption{Intensity profiles  across the dash lines  of  the first two samples in Fig. \ref{fig4}: first two rows show the intensity images and the   profiles of the first sample and last two  rows show  those for the second sample. The intensity profiles from the original image, the image processed by GIF and the image processed by the proposed method are shown in red dash, blue dotted and green solid lines, respectively. From left to right are the original image, image processed by GIF and image processed by the proposed method.}  \label{fig555}
\end{figure*}

 Visually, it is difficult to tell if GIF has oversmoothed some regions from the images directly.  In this paper, we show the intensity profiles to highlight the effect. Fig. \ref{fig555} shows two example where the intensity profiles of   horitonal lines (in white) are plotted. The intensity profiles from original image, the image processed by GIF and the image processed by the proposed SGRIF method are shown in red, blue, and green, respectively. From the second row, we can see  that both GIF and SGRIF improve the contrast near the vessels and SGRIF performs  slightly better. From the third row, we can hardly tell the location of  optic cup boudnary from the profiles in red and the blue while we observe a more obvious gradient change in the area pointed by the arrow in the profile in green. The small difference in the profiles can be critical for subsequent analysis tasks， as supported by the optic cup boundary dection results. 

%

\subsubsection{Sparse Learning based CDR Computation}
In the second test, we explore how the  decloud affects a direct vertical cup to disc ratio (CDR) assessment using sparse learning.  Previously, we have proposed a sparse dissimilarity-constrained coding (SDC) \cite{CJ15} algorithm to reconstruct each testing disc image $y$ with a set of $n$ training or reference disc
images $X=[x_1, x_2, \cdots, x_n$ with known CDRs $r=[r_1, r_2, \cdots, r_n]$.
The method is essentially a sparse learning based approach. 
SDC aims to find a solution $w$ to approximate $y$ by $Xw$ based on the following objective
function:
\begin{equation}
\arg\min_w \|y-Xw\|^2+\lambda_1\cdot \|d\circ w\|^2+\lambda_2 \cdot \|w\|_1,
\label{eqSDC}
\end{equation}
where $d=[d_1, d_2, \cdots, d_n]$ denote the similarity cost between $y$ and $X$, $\circ$ denotes dot product. 
The CDR of testing image $y$ is then estimated as
\begin{equation}
\hat{r}=\frac{1}{\textbf{1}^T_ww}r^Tw,\end{equation} 
where $\textbf{1}$ is a vector of 1s with length $n$.
More details of the SDC method are kept the same as that in \cite{CJ15} and can be found in the paper.

To justify the benefit of SGRIF for CDR computation, we
conduct CDR computation based on (\ref{eqSDC}). Similar to that in optic cup segmentation, we conduct tests in three scenarios: 1) original reference and testing images 2)  all reference and testing images are processed by GIF; 3) all reference and testing images are processed by SGRIF. 
 Table \ref{table3} shows the
CDR error  between automatically computed CDRs
and the manually measured CDRs in the testing images.  The CDR errors for retinal images of  eyes with and without cataract are computed separately. From the results, we obtain a relative reduction of 3.7\% for retinal images of eyes with cataract and a relative reduction of 2.4\% for retinal images of eyes without cataract.  It is worth to note that our method also improves the vertical CDR estimation even though it mainly improves the edges at the temporal side of the optic disc. This is not surprising as the temporal side is an important part in the reconstruction. In additional, we have also invited an ophthalmologist to manually measure a second set of CDRs \cite{CJ15} and  the inter-observer error is given in Table \ref{table3} as well.

\begin{table}
  \caption{ Accuracy of CDR Computation
} \begin{center}\
        \begin{tabular}{c|c|c| c    } \hline
        & All &   Cataract   & No   Cataract      \\\hline
            Original    & 0.0657 & 0.0626 &    0.0674       \\\hline
           GIF \cite{Hekaiming2013}  & 0.0665 & 0.0625  & 0.0686   \\ \hline
       \textbf{Proposed } & \textbf{0.0639}  &   \textbf{0.0603} &  \textbf{0.0658}     \\\hline
       Inter-Observer Error & 0.0767 & 0.0761& 0.0771 \\ \hline 
        \end{tabular}
    \end{center}
 \label{table3}
\end{table}

 \subsection{Performance on other regions}
\begin{figure*}
		\centering
		{\subfigure{
			\includegraphics[height=2in]{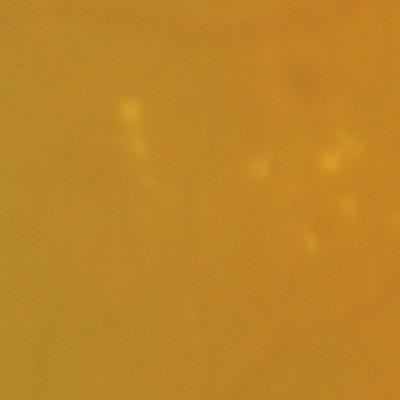}
	}} 
 	{\subfigure{
 			\includegraphics[height=2in]{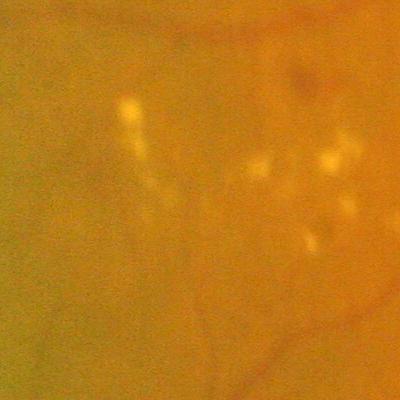}
 	}}
	{\subfigure{
			\includegraphics[height=2in]{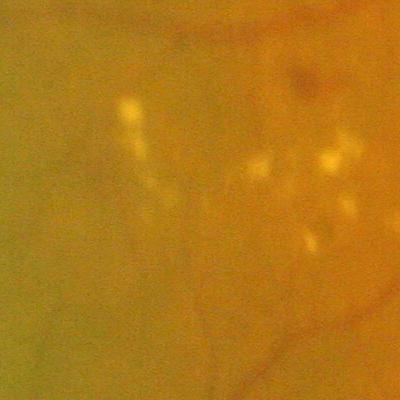}
	}}\\
	{\subfigure{
		\includegraphics[height=2in]{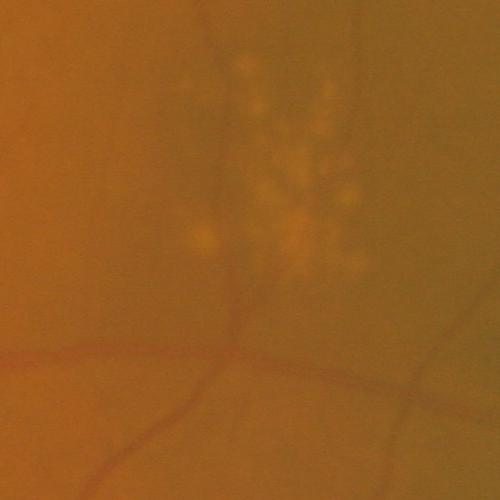}
}} 
{\subfigure{
		\includegraphics[height=2in]{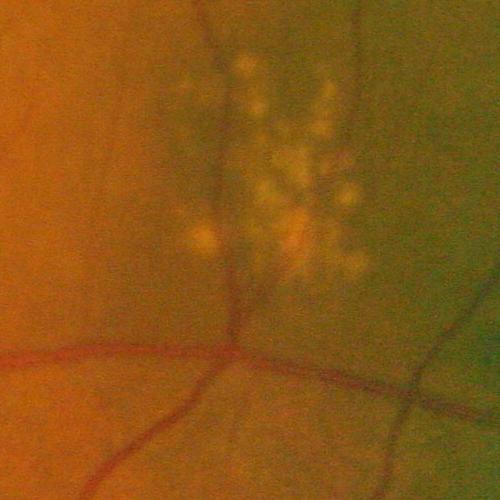}
}}
{\subfigure{
		\includegraphics[height=2in]{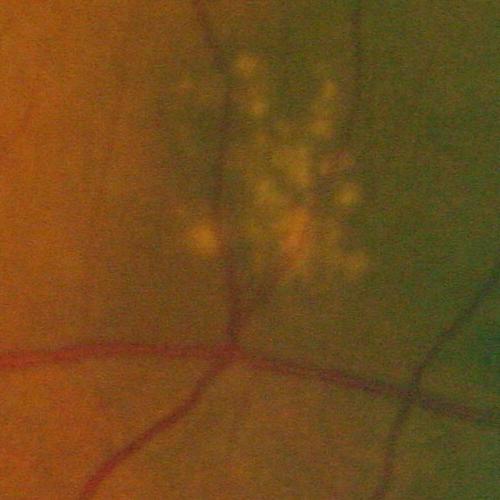}
}}\\
 	{\subfigure{
		\includegraphics[height=2in]{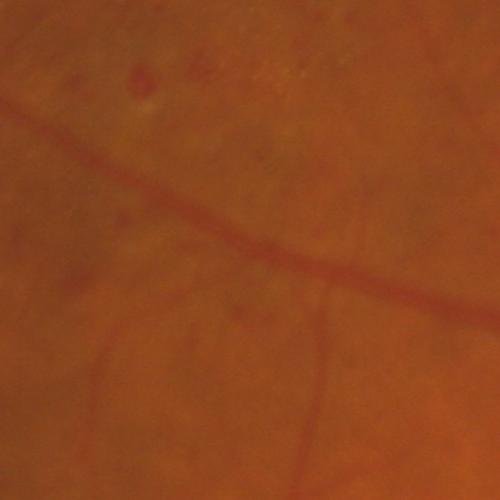}
    }} 
    {\subfigure{
		\includegraphics[height=2in]{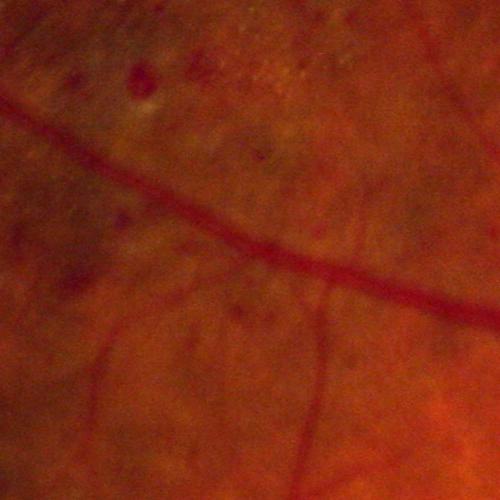}
    }}
    {\subfigure{
		\includegraphics[height=2in]{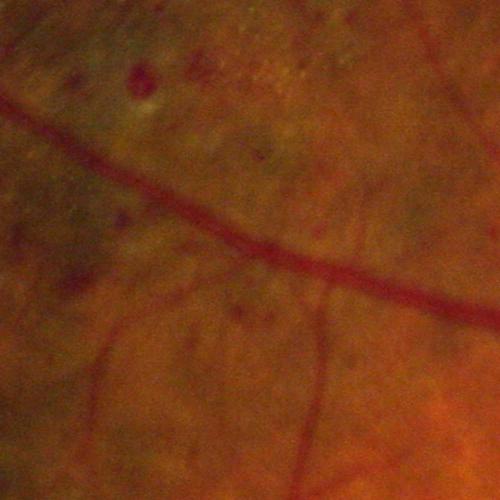}
    }}\\ \setcounter{subfigure}{0}
	{\subfigure[]{
		\includegraphics[height=2in]{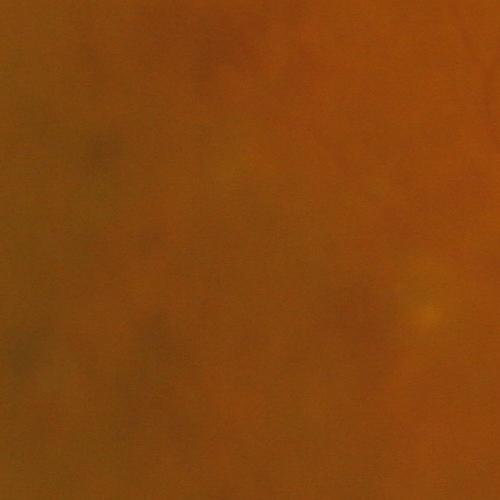}
}} 
{\subfigure[]{
		\includegraphics[height=2in]{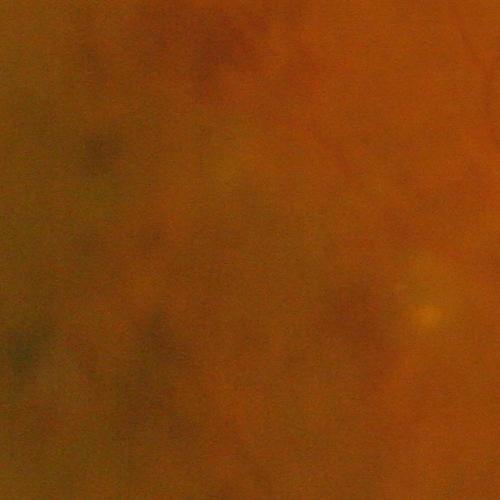}
}}
{\subfigure[]{
		\includegraphics[height=2in]{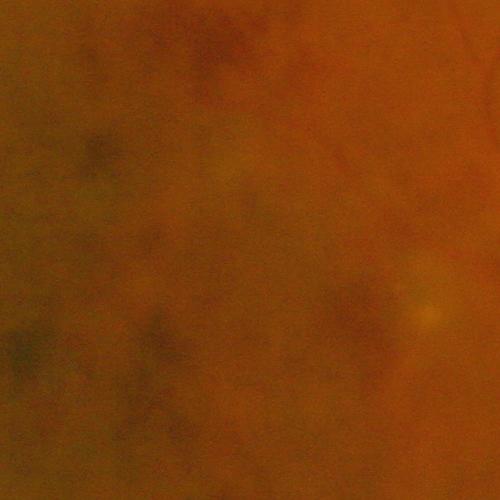}
}}

 \caption{Performance of the method in region with lesions:  (a) Original (b) image processed by GIF (c) image processed by SGRIF (best viewed in softcopy)}  \label{figlesions}
\end{figure*}

Besides the region around the optic disc, we have also applied our method to other regions including those with different lesions. Fig. \ref{figlesions} shows some results. From the results, we can see that both GIF and SGRIF improve the contrast of the original images.  Howerver, we do not have a  quantitative  measurement on  how GIF or SGRIF improves subsequent analysis in lesion area. 

 \section{Conclusions}
 Image quality is an important fact that affecting the performance of retinal image analysis. 
 In this paper, we propose  structure-preserving guided retinal image filtering to remove artifacts caused by a cloudy human-lens.   Our results show that the proposed method is able to improve the contrast within the retinal images, which is reflected by histogram flatness measurement, histogram spread and variability of local luminosity.  We further validate the algorithm with subsequent optic cup segmentation using deep learning and cup-to-disc ratio measurement using sparse learning. Both the experiments in optic cup segmentation and cup-to-disc ratio computation  show that the proposed  method  is beneficial for the subsequent tasks.
  In this paper, we mainly apply the algorithm in the region around the disc for optic disc analysis. We do not apply that for optic disc segmentation as the optic disc boundary is usually much stronger and will not be smoothed away in GIF. Therefore, our method does not improve the results.  Since our method improves the CDR measurement, it could potentially be applied for glaucoma detection. Our results also suggest that the method improve the blood vessel qualitatively.  In future work,  we will also evaluate how the processing affect other analysis such as vessel segmentation and lesion detection quantitatively. 
 
\bibliographystyle{IEEEbib}
\bibliography{bibdata}
\end{document}